\documentclass[lettersize,journal]{IEEEtran}
\usepackage{amsmath,amsfonts}
\usepackage{algorithmic}
\usepackage{algorithm}
\usepackage{array}
\usepackage[caption=false,font=normalsize,labelfont=sf,textfont=sf]{subfig}
\usepackage{textcomp}
\usepackage{stfloats}
\usepackage{url}
\usepackage{verbatim}
\usepackage{graphicx}
\usepackage{cite}

\usepackage{mwe}
\usepackage{makecell}
\usepackage{multirow}
\usepackage{bm}
\usepackage[table]{xcolor}
\usepackage{hyperref}
\usepackage{enumitem}
\usepackage{capt-of}
\usepackage{tabu}
\usepackage{booktabs}
\usepackage{tikz,xcolor}
\definecolor{lime}{HTML}{A6CE39}
\DeclareRobustCommand{\orcidicon}{%
	\begin{tikzpicture}
	\draw[lime, fill=lime] (0,0) 
	circle [radius=0.16] 
	node[white] {{\fontfamily{qag}\selectfont \tiny ID}};
	\draw[white, fill=white] (-0.0625,0.095) 
	circle [radius=0.007];
	\end{tikzpicture}
	\hspace{-2mm}
}

\foreach \x in {A, ..., Z}{%
	\expandafter\xdef\csname orcid\x\endcsname{\noexpand\href{https://orcid.org/\csname orcidauthor\x\endcsname}{\noexpand\orcidicon}}
}

\begin{document}

\newcolumntype{b}{>{\columncolor[RGB]{249, 248, 239}}c}  
\newcolumntype{d}{>{\columncolor[RGB]{230, 242, 255}}c}  
\newcolumntype{g}{>{\columncolor[RGB]{249, 239, 239}}c}  
\newcolumntype{e}{>{\columncolor[RGB]{234, 245, 246}}c}  
\newcolumntype{f}{>{\columncolor[RGB]{255, 239, 213}}c}  

\title{\vspace{-10pt}Refined Response Distillation for Class-Incremental Player Detection}

\author{Liang Bai\orcidA{}, Hangjie Yuan\orcidB{}, Tao Feng\orcidC{}, Hong Song\orcidD{}, Jian Yang\orcidE{}
\thanks{\textit{(Corresponding authors: Tao Feng; Hong Song.)}
}

\thanks{Liang Bai and Hong Song are with the School of Computer Science and Technology, Beijing Institute of Technology, Beijing 100081, China (e-mail: bai1911@foxmail.com; songhong@bit.edu.cn).}

\thanks{Hangjie Yuan is with the College of Control Science and Engineering, Zhejiang University, Hangzhou 310027, China (e-mail: hj.yuan@zju.edu.cn).}

\thanks{Tao Feng is with the College of Computer Science, Sichuan University, Chengdu 610065, China (e-mail: fengtao@stu.scu.edu.cn).}

\thanks{Jian Yang is with the School of Optics and Photonics, Beijing Institute of Technology, Beijing 100081, China (e-mail: jyang@bit.edu.cn).}

}

\markboth{Journal of \LaTeX\ Class Files,~Vol.~14, No.~8, August~2021}%
{Shell \MakeLowercase{\textit{et al.}}: A Sample Article Using IEEEtran.cls for IEEE Journals}


\maketitle

\begin{abstract}

Detecting players from sports broadcast videos is essential for intelligent event analysis. 
However, existing methods assume fixed player categories, incapably
accommodating the real-world scenarios where categories continue to evolve. 
Directly fine-tuning these methods on newly emerging categories also exist the catastrophic forgetting due to the non-stationary distribution. 
Inspired by recent research on incremental object detection (IOD), we propose a \textbf{Refined Response Distillation} ($R^2D$) method to effectively mitigate catastrophic forgetting for IOD tasks of the players. Firstly, we design a progressive coarse-to-fine distillation region dividing scheme, separating high-value and low-value regions from classification and regression responses for precise and fine-grained regional knowledge distillation. Subsequently, a tailored refined distillation strategy is developed on regions with varying significance to address the performance limitations posed by pronounced feature homogeneity in the IOD tasks of the players. 
Furthermore, we present the NBA-IOD and Volleyball-IOD datasets as the benchmark and investigate the IOD tasks of the players systematically. 
Extensive experiments conducted on benchmarks demonstrate that our method achieves state-of-the-art results.
The code and datasets are available at \url{https://github.com/beiyan1911/Players-IOD}.

\end{abstract}


\section{Introduction}

\IEEEPARstart{T}{he} advancements in computer vision have led to numerous studies~\cite{refx1,refx2,refx3,refx4,refx5,refx6,refx7,refx8,refx9,refxb9,refx10,refx11,refxb11,refxc11} aimed at enhancing the intelligent analysis of sports videos. Here, our research focuses on player detection in broadcast videos. Existing approaches~\cite{refx3,refx4,refx7,refx10} primarily concentrate on constructing optimal detection patterns derived from ground truth data. However, these methods tend to neglect the challenges posed by incremental learning due to the dynamic nature of data flow in real-world sports scenarios. As the categories of players continuously evolve across a series of sports competitions (\textit{e.g.}, player transfers, new recruits, etc.), training a detector after collecting and annotating all broadcast videos following traditional routes which assume fixed object categories would result in significant time delays and increased workload. For instance, in the NBA, a season typically involves dozens of participating teams and multiple rounds of matches between them, leading to a non-stationary data distribution. 
Furthermore, training an effective player detection model from scratch for base and novel sets of players is often impractical due to privacy concerns, \textit{i.e.}, parts of the data might not be available when we target retraining a model.

\begin{figure}[!t]
\centering
\includegraphics[width=0.42\textwidth]{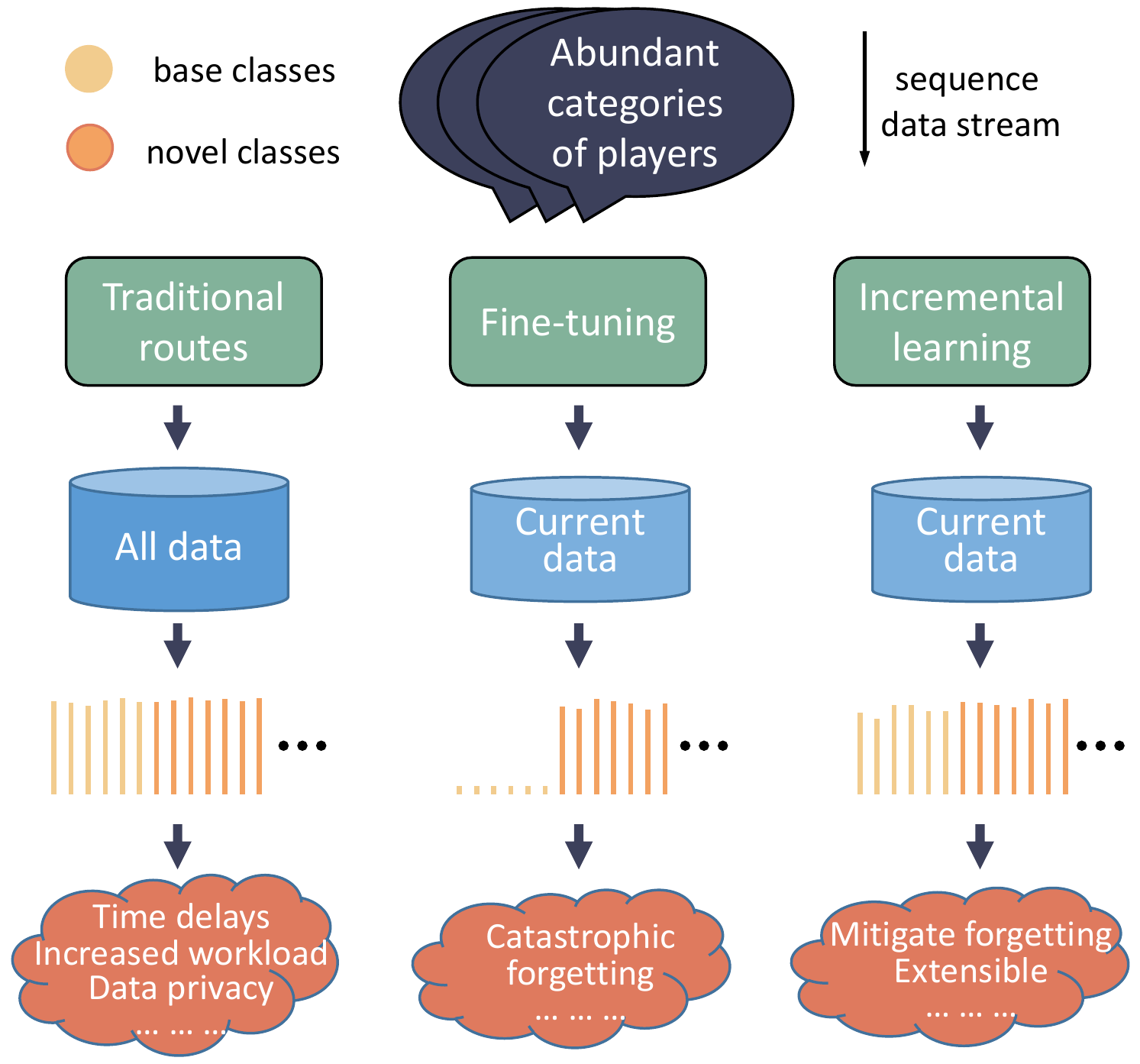}\vspace{-1em}
\caption{\small Differences in traditional routes, fine-tuning, and
incremental learning in detecting abundant categories of players.}
\label{fig_forget}
\vspace{-12pt}
\end{figure}

A naive approach entails fine-tuning the detector to accommodate the continuously changed task. Nevertheless, directly fine-tuning leads to the forgetting of base classes while learning novel classes, as a result of the dynamic data distributions. Figure \ref{fig_forget} highlights the significant disparities in player detection across various approaches. 
From this observation, it is evident that directly fine-tuning results in catastrophic forgetting, while incremental learning retains much of previous knowledge. In this work, we aim to address the \textbf{incremental detection of players in broadcast videos}. Specifically, we propose a tailored incremental detection method for class incremental learning, targeting player detection in sports broadcast scenarios.

Class incremental object detection (IOD) aims to adapt to new classes without compromising previously acquired knowledge. 
Several studies~\cite{refx12,refx13,refx14,refx15,refx16,refx17,refx18} employ knowledge distillation (KD), circumventing privacy concerns, to mitigate the issue of forgetting. 
However, these works primarily target general static scenes, such as MS COCO~\cite{refx19}, and often exhibit performance limitations when directly applied to real-world sports events, due to the absence of a customized incremental learning strategy.
Sports broadcast scenes for incremental player detection pose greater complexity. 
Despite the rapid movement, pose changes, and occlusions involving players, the primary challenge stems from the pronounced feature homogeneity resulting from identical team uniforms.
As a consequence, knowledge transfer in incremental learning becomes more challenging in such scenarios.
Presently, the incremental detection of players remains an under-explored area.

To effectively address the issue of feature homogeneity, this paper revisits the IOD tasks associated with player detection in sports broadcast scenes.
General methods to solve IOD usually ignore this critical issue.
For instance, ERD~\cite{refx17} preserves knowledge solely from high-certainty regions, neglecting the potential benefits of low-value regions in discriminating players.
Object detection distillation methods~\cite{refx20, refx21} also fail to consider this issue.
For instance, Guo \textit{et al.}~\cite{refx20} decouple the features of object and non-object regions, yielding enhanced knowledge transfer performance in feature imitation.
Similarly, Zheng \textit{et al.}~\cite{refx21} employ classification and localization distillations in distinct regions, resulting in improved distillation performance in logit mimicking.
Nonetheless, these methods~\cite{refx20, refx21} rely on a standard reference (the ground truth) to formulate their distillation strategies, which is not available in incremental scenarios.
We empirically observe that utilizing fine-grained regions for distillation facilitates the detection of players, since these regions include informative details to discriminate players.
Inspired by this, we propose \textbf{R}efined \textbf{R}esponse \textbf{D}istillation ($R^2D$) that progressively divides regions from coarse to fine for precise and fine-grained regional knowledge distillation.

To be more specific, $R^2D$  consists of two essential steps.
Initially, we select coarse distillation candidate regions based on classification response threshold, drawing partial inspiration from post-processing operations that filter noise and background through detection score threshold in standard detectors such as RetinaNet \cite{refx22}, GFL \cite{refx23}, AutoAssign \cite{refx24}, and YOLOX \cite{refx25}.
Subsequently, we segregate high-value and low-value regions from the common candidate regions using a clustering method with distillation quality indicators.
Considering the inconsistency of the features in the classification and regression branches \cite{refx26,refx27,refx28}, we allocate refined distillation regions on both heads simultaneously to achieve efficient knowledge retention. 
In doing so, we introduce a more refined response distillation scheme for preserving more fine-grained knowledge in incremental player detection within real-world scenarios. 
For the classification head, we integrate a specifically designed response decoupling loss and L1 loss for the high-value and low-value regions. 
For the regression head, we implement localization distillations (LD) with distinct temperatures for the high-value and low-value regions.
Additionally, we produce the NBA-IOD and Volleyball-IOD datasets as the benchmark of incremental player detection. 
In this work, we systematically investigate the IOD task of players from real-world scenarios and perform extensive experiments to support our analysis and conclusion. The contributions of our work can be summarized as follows:

\begin{enumerate}[label=\arabic*)]
	\item To the best of our knowledge, this paper is the first work to systematically investigate the IOD tasks of players.
	\item We propose the $R^2D$ method, which performs refined distillation for regions with different values, to more efficiently mitigate the catastrophic forgetting for IOD tasks of the players.
	\item We construct two datasets NBA-IOD and Volleyball-IOD as the benchmark, providing data support for IOD tasks of the players.
	\item Extensive experiments on two benchmark datasets demonstrate that our proposed method achieves state-of-the-art performance.
\end{enumerate}

\section{Related Works}

\subsection{Player Detection and Identification}
Detecting and identifying players in sports broadcast videos is vital for the intelligent analysis of sports events. Recently, Acuna \cite{refx3} \textit{et al.} proposed a real-time framework to detect and track players in basketball game videos. Buric \textit{et al.} \cite{refx4} overviewed the performance of several standard detectors in the handball scenario. Laine \textit{et al.} \cite{refx5} designed a self-supervised pipeline to detect and track soccer players from low-resolution scenarios under different recording conditions. Usha Kiruthika \textit{et al.} \cite{refx8} analyzed the performance of several object detection approaches in the soccer scenario. Feng \textit{et al.} \cite{refx9} used a graph-based module to dynamically capture the person representation of the players and achieved better player identification accuracy in sports videos. Zheng \textit{et al.} \cite{refx10} explored a lightweight detector YOLO-OSA to detect players in moving videos as a guide for systematic training. However, these methods are scarce in the exploration of incremental learning. In this paper, we systematically investigate the IOD task of players from real-world scenarios.

\subsection{Incremental Object Detection}
Traditional object detectors forget learned base sets of categories when learning novel sets of categories due to non-stationary data distribution. To tackle this issue, a series of works are proposed in IOD\cite{refx12,refx13,refx14,refx15,refx16,refx17,refx18,refx29,refx30,refx31,refx32}. Shmelkov \textit{et al.} \cite{refx12} first applied LwF \cite{refx33} in IOD and proposed an incremental detector. Li \textit{et al.} \cite{refx13} proposed a real-time and efficient incremental detector RILOD on edge devices. Peng \textit{et al.} \cite{refx14} proposed a Selective and Inter-related Distillation (SID) to improve the incremental learning capabilities of anchor-free detectors. Feng \textit{et al.} \cite{refx17} proposed an elastic strategy to learn the responses of the classification head and regression head and remarkably improved the performance of the IOD tasks. Joseph \textit{et al.} \cite{refx31} leveraged meta-learning to optimize sharing information for minimizing forgetting of the incremental detection. Zhao \textit{et al.} \cite{refx32} proposed a 3D domain-adaptive class incremental object detection framework to overcome forgetting. In this paper, we explore a refined method to handle incremental detection issues and apply it to solve the real-world problem of class incremental detection in sports videos.

\subsection{Knowledge Distillation in Object Detection}
KD is a common approach to transferring knowledge in object detection. Hinton \textit{et al.} \cite{refx34} first proposed KD to transfer knowledge from complex networks to compact networks. Chen \textit{et al.} \cite{refx35} first introduced KD in object detection to get a smaller and faster object detector. Subsequently, an increasing number of studies~\cite{refx20,refx36,refx21,refx37,refx38,refx39} are devoted to this line. Zheng \textit{et al.} \cite{refx21} re-formulated the KD process of localization information in object detection, and achieved better distillation performance on logit imitation. Dai \textit{et al.}\cite{refx36} took feature, response, and relation into the KD process to treat the neglect of valuable relational information between instances. Ma \textit{et al.} \cite{refx38} proposed a hierarchical visual-language KD method for one-stage detection. Yang \textit{et al.} \cite{refx39} proposed focal and global distillation for object detection to address the differences between teacher and student detectors. Furthermore, KD is also introduced to mitigate catastrophic forgetting in IOD~\cite{refx17,refx13,refx12,refx31}. In this paper, we propose a refined distillation method from coarse to fine for precise regional knowledge distillation.

\begin{figure*}
\centering
\includegraphics[width=\textwidth]{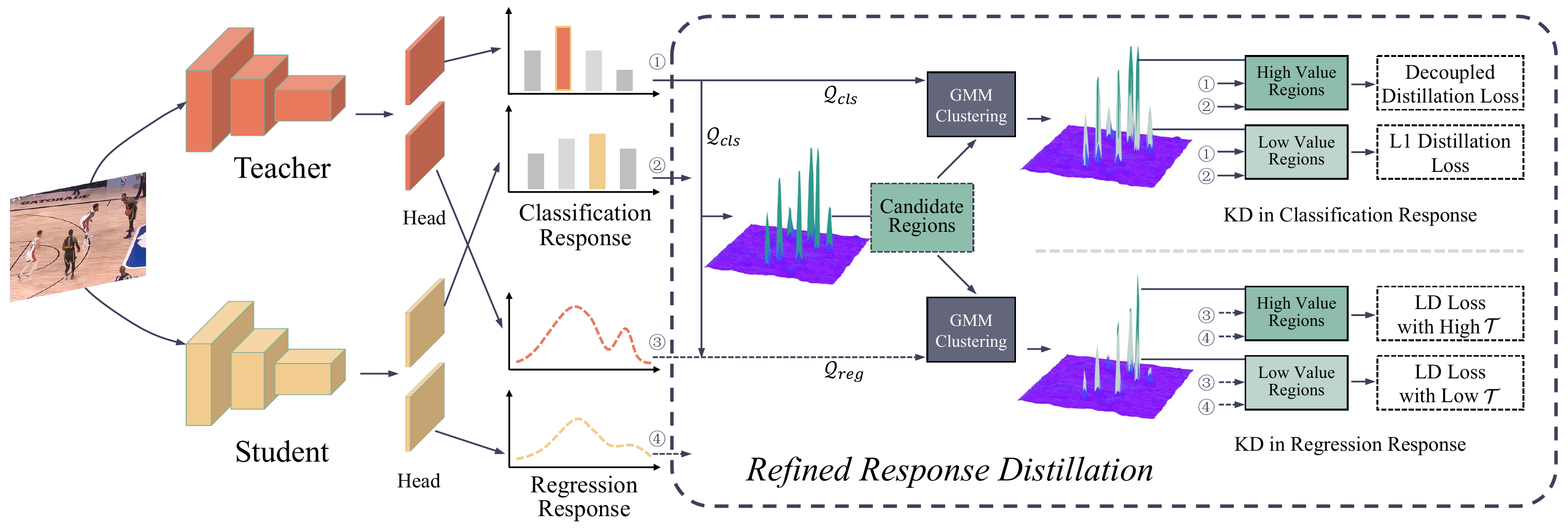}\vspace{-1em}
\centering
\caption{\small The overall architecture of refined response distillation for class incremental detection of players. 
}
\label{fig_structure}
\end{figure*}

\section{Proposed Method}

\subsection{Overview and Formulation}
The IOD tasks of the players in sports broadcast scenes require a customized distillation scheme to more effectively mitigate catastrophic forgetting, tackling the pronounced feature homogeneity. 
In this paper, we employ GFL~\cite{refx23} as both teacher and student detectors and accomplish this objective through the proposed refined response incremental distillation $R^2D$.
Figure \ref{fig_structure} shows the overall framework of the proposed $R^2D$. Firstly, we define response quality indicators for the classification and regression heads separately and preliminarily determine the common candidate regions on both heads. Then, to fully exploit the candidate regions for efficiently transferring knowledge, we use a clustering method to segregate high-value and low-value regions on both heads. Finally, we construct a tailored knowledge distillation strategy for the various valuable regions.

Let $P$ denote all categories of players that are incrementally introduced to the detector, $P_b$ denote the base sets of categories, and $P_n$ denote the novel sets of categories. Players’ IOD aims to learn a detector that can incrementally detect novel classes $P_n$ without forgetting the previous classes $P_b$. Let $T$ represent the teacher detector (learned in the base classes), $S$ represent the student detector (incrementally learning in the novel classes), $\mathcal{C}_{T}$ and $\mathcal{B}_{T}$ are the responses of classification head and regression head in teacher detector. The overall learning objective of the student detector is defined as:

\begin{equation}
\label{equation_total_loss}
\begin{aligned}
   \mathcal{L}_{total}=\mathcal{L}_{detector}
+\lambda _{1}\mathcal{L}_{cls\_distill}
+\lambda _{2}\mathcal{L}_{reg\_distill} 
\end{aligned}
\end{equation}
where $\mathcal{L}_{detector}$ is the classification and localization loss for learning novel categories. $\mathcal{L}_{cls\_distill}$ and $\mathcal{L}_{reg\_distill}$ are the refined distillation losses for classification and regression heads respectively, which are used to retain the knowledge of the base classes. $\lambda_{1}$ and $\lambda_{2}$ are hyper-parameters to balance the weight of different parts. We use $\lambda_{1}=\lambda_{2}=1$ by default.

In the following sections, we will elaborate on the detailed process of $R^2D$. 

\subsection{Dividing Distillation Regions From Coarse to Fine}\label{sec_method_div}
\textbf{Rough Candidate Regions.} 
Since box annotations of base classes are not available in the data of new classes, it is challenging to distinguish valuable regions of the response from non-valuable ones. 
ERD \cite{refx17} overcomes this challenge by dynamically selecting high-certainty regions from the maximum response to retain knowledge while neglecting other potential benefits of low-value regions. 
Inspired by this selection paradigm, our objective is to develop a coarse-to-fine approach that optimizes region selection for distillation.
First, we employ a straightforward thresholding technique to instantiate a rough selection process.

In standard detectors, removing background and noise regions by applying a threshold (usually 0.05) on detection confidence is a default post-processing technique.
Drawing inspiration from this practice, denoting one feature vector on the response of the classification head or regression head as a node, we calculate the distillation quality (confidence) of each node in the classification response as follows:
\begin{equation}
\label{equation_cls_quality}
\mathcal{Q}_{cls}={\rm max}(\sigma(\mathcal{C}_{T}))
\end{equation}
where $\sigma$ represents the ${\rm Sigmoid}$ activation function and $\rm max$ represents the maximum confidence of each response node in classification head. Then, formulate the rough distillation candidate regions by imposing a threshold $\theta$ on the distillation quality of classification response as follows:
\begin{equation}
\label{equation_cand_region}
R_{cand}=\mathcal{Q}_{cls}>\theta
\end{equation}
where $\theta$ determines regions with different sizes. We keep the same as the traditional setting of previous work, select $\theta = 0.05$, and verify its effectiveness through experiments in Table~\ref{table_abla_thela}.

On the response of the regression head, each edge of the bounding box is re-expressed as a general distribution following GFL\cite{refx23}, and the bounding box in per response node is defined as $\mathcal{G}=\left [p_{t},p_{b},p_{l},p_{r}\right ]\epsilon \mathbb{R}^{n\times 4}$. ERD\cite{refx17} takes the largest element among the distributions as the localization confidence of response nodes. We argue that the localization confidence is not only related to the classification quality (if it is too low, the proposal will be eliminated and cannot be optimized) but also related to the information entropy of the general distributions of bounding boxes. Thus, the distillation quality in the regression head is defined as follows:

\begin{equation}
\label{equation_reg_quality}
\mathcal{Q}_{reg}
=-log(1-\mathcal{Q}_{cls})
*{\rm max}(E(p)\mid p\epsilon \mathcal{G})
\end{equation}

\begin{equation}
\label{equation_information_entroy}
E(p)=\sum_{i=1}^{M}- {p}^{i}*\log_{2} {p}^{i}
\end{equation}
where $E(p)$ is the information entropy for one general distribution $p$, and $M$ is the number of elements in each general distribution.

\begin{figure}[!t]
\centering
\includegraphics[width=0.48\textwidth]{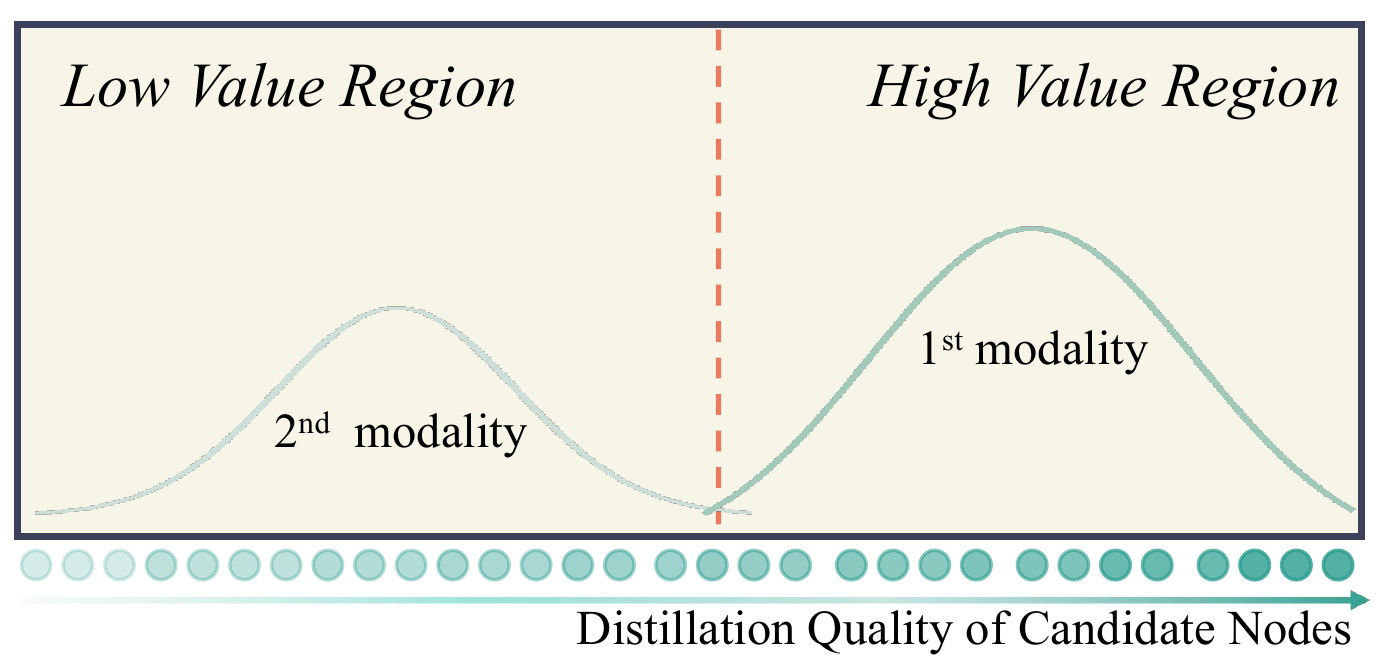}\vspace{-1em}
\caption{\small Segregating high-value and low-value regions from the rough candidate regions according to distillation quality using the GMM with two modalities.}
\label{fig_gmm}\vspace{-12pt}
\end{figure}

\textbf{Refined Distillation Regions.}
A flexible way to refine the candidate regions into different parts is clustering. 
Inspired by work \cite{refx40}, we model the distillation quality of the candidate regions in classification and regression heads as Gaussian Mixture Model (GMM) with two modalities—one indicates the space of the high-value regions, and the other indicates the space of the low-value regions.
Given a one-dimensional classification or regression quality values, we optimize the GMM parameters by the Expectation Maximization algorithm to determine the probability that each response node of the candidate regions is in the high-value or low-value regions and then dynamically divide all the candidate regions into two clusters. 
Figure \ref{fig_gmm} illustrates the schematic diagram using GMM to refine the rough candidate regions. 
The high-value and low-value regions in the classification head are divided as follows:
\begin{equation}
\label{equation_cls_regions}
R_{cls}^{H}, R_{cls}^{L}=GMM(\mathcal{Q}_{cls}, R_{cand})
\end{equation}

The high-value and low-value regions in the regression head are divided as follows:
\begin{equation}
\label{equation_reg_regions}
  R_{reg}^{H},R_{reg}^{L}=GMM(\mathcal{Q}_{reg},R_{cand})  
\end{equation}

According to the refined regions $R_{cls}^{H}$, $R_{cls}^{L}$, $R_{reg}^{H}$, $R_{reg}^{L}$, we extract the corresponding nodes $\mathcal{C}_{T}^{H}$, $\mathcal{C}_{T}^{L}$, $\mathcal{B}_{T}^{H{}'}$, $\mathcal{B}_{T}^{L{}'}$ from the classification response $\mathcal{C}_T$ and the regression response $\mathcal{B}_T$. Then perform Non-Maximum Suppression (NMS) on the segregated response nodes in the regression head as follows:

\begin{equation}
\label{equation_reg_regions}
\mathcal{B}_{T}^{H}={\rm NMS}( \mathcal{B}_{T}^{H{}'})
\end{equation}
\begin{equation}
\label{equation_reg_regions}
\mathcal{B}_{T}^{L}={\rm NMS}( \mathcal{B}_{T}^{L{}'})
\end{equation}

The next part will show the tailored distillation strategy on $\mathcal{C}_{T}^{H}$, $\mathcal{C}_{T}^{L}$, $\mathcal{B}_{T}^{H}$ and $\mathcal{B}_{T}^{L}$.

\subsection{Tailored Distillation Strategy for Different Regions}
{\bf {Refined Classification Distillation.}} 
Regions exhibiting higher classification quality $\mathcal{Q}_{cls}$ are more likely to contain targets, potentially leading to a more pronounced presence of knowledge relating to target categories.
Inspired by DKD \cite{refx41}, we decouple the classification KD in the high-value regions $R_{cls}^{H}$ into maximum class KD $\mathcal{L}_{max\_cls}^{H}$ and non-maximum class KD $\mathcal{L}_{not\_max\_cls}^{H}$ to circumvent their mutual constraints. 
To separate the response elements relevant and irrelevant to the maximum class, we define $ b^{H} =[p_{max}^{H},p_{not\_max}^{H}]\epsilon \mathbb{R}^{1\times 2}$, which represents the binary probabilities of the maximum class $p_{max}$ and all the other non-maximum class $p_{not\_max}$ in $R_{cls}^{H}$.
Denoting the classiﬁcation probabilities of one sample using $\mathcal{Z}=[z_{1},z_{2},...,z_{N}]\epsilon \mathbb{R}^{1\times N}$, they can be calculated as follows: 

\begin{equation}
\label{equation_cls_max}
p_{max} =\frac{ {\rm exp} ({\rm max}(\mathcal{Z}))}{\sum_{i=1}^{N}{\rm exp}(z_{i})}
\end{equation}

\begin{equation}
\label{equation_cls_not_max}
p_{not\_max}=\frac  {\sum_{j=1,j\ne {\rm max\_id}(\mathcal{Z}) }^{N}{\rm exp}(z_{j})}   {\sum_{i=1}^{N}{\rm exp}(z_{i})}
\end{equation}
where ${\rm max}(\mathcal{Z})$ and ${\rm max\_id}(\mathcal{Z})$ represent the largest element of $\mathcal{Z}$ and its index respectively, and $N$ is the number of player categories.
We use Kullback-Leibler divergence as the knowledge distillation loss of the maximum class:
\begin{equation}
\label{equation_cls_loss_max}
\mathcal{L}_{max\_cls}^{H}=KL(b_{T}^{H}\parallel b_{S}^{H})
\end{equation}
Let $\hat{P}=[\hat{p}_{1},...,\hat{p}_{m-1},\hat{p}_{m+1},...,\hat{p}_{N}]\epsilon \mathbb{R}^{1\times (N-1)}$ denote the probabilities of non-maximum elements of each response node in $R_{cls}^{H}$, where:
\begin{equation}
\label{equation_p_j}
\hat{p}_{j}
=\frac{{\rm exp}{(z_{j})}}
{ {\textstyle \sum_{i=1,i\ne {\rm max\_id}(\mathcal{Z}) }^{N}
{\rm exp}(z_{i})
} }
\end{equation}
The knowledge distillation loss of the non-maximum class in $R_{cls}^{H}$ is:
\begin{equation}
\label{equation_cls_loss_not_max}
\mathcal{L}_{not\_max\_cls}^{H}
=KL(\hat {P}_{T}^{H}\parallel \hat {P}_{S}^{H})
\end{equation}
Then, the distillation loss for the high-value regions of the classification head is defined as:
\begin{equation}
\label{equation_cls_loss_high}
\mathcal{L}_{cls}^{H}
=\lambda_{3}\mathcal{L}_{max\_cls}^{H}
+ \lambda_{4}\mathcal{L}_{not\_max\_cls}^{H}
\end{equation}
where $\lambda_{3}$ and $\lambda_{4}$ are hyper-parameters to balance the weight of different parts, and we use $\lambda_{3}=\lambda_{4}=1$ by default. 
It should be noted that our decouple method is different from DKD \cite{refx41}, which decouples classification KD into target class KD and non-target class KD according to the ground truth to independently adjust their importance. 
In IOD tasks of the players, we decouple the classification KD of the high-value regions into maximum class KD and non-maximum classes KD according to the classification response to reduce mutual inhibition of noticeable response differences.

The low-value regions of the classification head is more likely to be the background in the image. In incremental learning, knowledge distillation in background regions is equally important. Therefore, we use the L1 norm  as the quick optimization goal to retain the partial background knowledge:
\begin{equation}
\label{equation_cls_loss_low}
\mathcal{L}_{cls}^{L}
=\left \| \mathcal{C}_{T}^{L}
-\mathcal{C}_{S}^{L}
\right \|
\end{equation}
Finally, the total distillation loss for the classification head is as follows:
\begin{equation}
\label{equation_cls_loss}
\mathcal{L}_{cls\_distill}=
\alpha  * \mathcal{L}_{cls}^{H}
+\beta * \mathcal{L}_{cls}^{L}
\end{equation}
where $\alpha$ and $\beta$ are the ratios of nodes in high-value and low-value regions to the number of nodes in the candidate regions, and the sum of $\alpha$ and $\beta$ is 1.

{\bf {Refined Localization Distillation.}} 
Different regions of response in the regression head have diverse characteristics.
Response from the high-value regions possesses higher certainty, while response from the low-value regions contains more noise over the incremental scenes. 
So, we use localization distillation with varying temperatures for high-value and low-value regions to fully exploit the candidate regions in the regression head. 
Using a higher distillation temperature in $R_{reg}^{H}$ will take care of every element of the general distribution in bounding box re-representation. 
Using a slightly lower distillation temperature in $R_{reg}^{L}$ can make the knowledge transfer in incremental learning pay more attention to the higher elements in the general distributions, avoiding the noise interference of the lower elements. 
The localization distillation loss in $R_{reg}^{H}$  is as follows:
\begin{equation}
\label{equation_reg_loss_high}
\mathcal{L}_{reg}^{H}
=KL 
(\mathcal{B}_{T}^{H}/\mathcal{T}_{1},
\mathcal{B}_{S}^{H}/\mathcal{T}_{1})
\end{equation}
The localization distillation loss in $R_{reg}^{L}$ is as follows:
\begin{equation}
\label{equation_reg_loss_low}
\mathcal{L}_{reg}^{L}
=KL 
(\mathcal{B}_{T}^{L}/\mathcal{T}_{2},
\mathcal{B}_{S}^{L}/\mathcal{T}_{2})
\end{equation}
Finally, The total loss of localization distillation is as follows:
\begin{equation}
\label{equation_reg_loss}
\mathcal{L}_{reg\_distill}=
\lambda_{5}\mathcal{L}_{reg}^{H}
+\lambda_{6}\mathcal{L}_{reg}^{L}
\end{equation}
where $\mathcal{T}_{1}$ and $\mathcal{T}_{2}$ are the temperatures of localization distillation. We take $\mathcal{T}_{1}=10$ consistent with LD\cite{refx21} and ERD\cite{refx17}. To ensure that $\mathcal{T}_{1} < \mathcal{T}_{2}$, we set $\mathcal{T}_{2}=5$ and also conduct experiments in Table~\ref{table_abla_t2} to verify its robustness. $\lambda_{5}$ and $\lambda_{6}$ are hyper-parameters to balance the weight of different parts, we use $\lambda_{5}=\lambda_{6}=1$ by default.

\section{Datasets and Scenario Settings}
\subsection{NBA-IOD}

We introduce the NBA-IOD dataset, a collection of images sampled from NBA games videos, specifically derived from broadcast videos of six games played during the 2019-2022 Finals. 
The source videos encompass two resolutions, $1920 \times 1080$ and $1280 \times 720$.
We randomly sample images from the broadcast footage and annotate the players present in these images.
The NBA-IOD dataset comprises 6 basketball teams, 50 players, 2,290 images, and 21,615 bounding boxes, averaging 432 bounding boxes per player.
We utilize images from the first half of the games for training and reserve those from the second half for testing purposes.

\textbf{Scenario settings}. 
Contrary to static scenes found in datasets such as MS COCO~\cite{refx19}, large-scale sports matches' broadcast videos, like those of the NBA, typically present different teams as a continuous stream. 
Players from the same team often appear in the same or adjacent shots. 
Consequently, dividing the abundant categories into incremental steps uniformly (\textit{e.g.}, class splits of 20+10+10+10 in COCO) may result in incremental splits that deviate from the real-world evolution of stream data, as detected players within the same incremental step could belong to different teams with varying match times.
Furthermore, professional sports event analyses frequently focus on the attributes of specific groups, distinct teams, home-and-away sides, and other scenes. Taking these factors into account, we have devised various scenarios that aim to include players from the same team in the same incremental step as much as possible. 
The subsequent divisions are as follows:

\begin{itemize}[leftmargin=*,topsep=0pt, noitemsep]

\item \textbf{Three-step setting}. 
We partition the NBA-IOD dataset into three tasks, with class distributions of 15+17+18.
In the three-step setting, we designate 15 players as base classes, and subsequently add 17 and 18 new players as novel classes in each step. 
To maintain consistency with the realistic class incremental scenario present in broadcast videos, we endeavor to allocate players from the same team to the same incremental step.

\item \textbf{Five-step setting}. 
To further test the performance of our method on a long sequence of tasks, we split the NBA-IOD into 5 tasks with class splits of 15+9+8+10+8. 
The five-step setting uses 15 players as base classes, and 9/8/10/8 new players are added each time as novel classes.

\item \textbf{Home-and-away setting}.
A prevalent class incremental learning scenario for players in real-world applications originates from the home-and-away match system. 
This system inherently introduces a class incremental challenge due to the alternation between home and away matches. 
Consequently, we employ game 5 of the finals in the 2019-2020 NBA playoffs from the NBA-IOD dataset to evaluate the performance of our method within the home-and-away setting. 
Specifically, this setting is structured as a two-step scenario, designating 9 players from the home team as base classes and 9 players from the away team as novel classes.

\item \textbf{Two-team setting}. 

To thoroughly examine the performance of our method in more complex scenarios, we present a distinct two-step setting that extends beyond the conventional home-and-away setting. 
In this configuration, each step of the IOD tasks encompasses both home and away situations, spanning multiple rounds of the games. 
The two-team setting designates nine players from Miami Heat as base classes, and nine players from Los Angeles Lakers as novel classes.

\end{itemize}

\subsection{Volleyball-IOD} 
Volleyball-IOD is collected from the full-court record of Serbia versus Italy in the quarter-final of the 2020 Tokyo Olympic women's volleyball games. Similar to the NBA-IOD dataset, we randomly sample images from the broadcast view and label the players in them. The Volleyball-IOD contains 2 teams, 14 players, 195 images, 2,025 objects, with an average of 144 images per identity. We use images in the first two rounds of the games for training and the last round for testing.

\textbf{Scenario setting}. 
Similar to the NBA-IOD dataset, we split the Volleyball-IOD into 2 tasks with the classes of 7+7. This setting uses 7 players from the Serbian team as base classes, and 7 players from the Italian team as novel classes. We follow the multi-step setting of NBA-IOD to guarantee consistency with incremental player detection of the real world.

\section{Experiments and Discussions}
In this section, we conduct experiments and analyses on various scenes from the NBA-IOD and Volleyball-IOD datasets to validate the proposed method. 
Subsequently, we perform ablation studies to demonstrate the effectiveness of each component. 
Lastly, we discuss the impact of hyper-parameters and further perspective on refined regions and superior performance.

{\bf{Experimental setting.}} 
We implement incremental detectors using MMDetection \cite{refx42} and compare it with several classical methods, such as RILOD\cite{refx13}, SID\cite{refx14}, and ERD\cite{refx17}. 
To ensure a fair comparison, we implement these methods on the GFL\cite{refx23} detector.
We reproduce RILOD following its original implementation details and reproduce the most improved version of SID as presented in its paper, since they are not publicly available.
For ERD, we use the official implementation. 
By default, training is performed on two NVIDIA RTX 3080Ti GPUs.
Each detector is trained for 12 epochs, with a batch size of 2 and an initial learning rate of 0.05.
Performance metrics, including $AP$, $AP_{50}$, and $AP_{75}$, are employed to evaluate all methods.

\subsection{Quantitative Results}
{\bf{ Three-step setting.}} 
We report the incremental results under the three-step setting to evaluate the performance of the proposed method. 
In Table~\ref{table_three}, ``Upper Bound'' denotes the result obtained by training on all data, while ``Catastrophic Forgetting'' denotes fine-tuning directly on new data in the incremental step, using parameters from the last step. 
Compared with $AP$ of 76.0\% in step 1, directly fine-tuning drops sharply to 41.9\% and 38.3\% in step 2 and step 3. 
The proposed $R^2D$ method substantially outperforms directly fine-tuning in terms of $AP$, $AP_{50}$, and $AP_{75}$ across all incremental steps.
This outcome stems from the inconsistent class distributions in streaming data.
As the detector learns novel player classes, the knowledge previously acquired from the base classes is disrupted, leading to catastrophic forgetting.
However, $R^2D$ mitigates this issue through the refined response incremental distillation strategy.
Moreover, the proposed method demonstrates notable improvement compared to RILOD, SID, and ERD in each incremental step and across all evaluation metrics, approaching the Upper Bound.
These results highlight the robustness of $R^2D$ in combating forgetting and its superior adaptability to player IOD tasks in broadcast scenarios.

\begin{figure}[!t]
\centering
\subfloat[\small Average Precision in Step 2]{\includegraphics[width=0.45\textwidth]{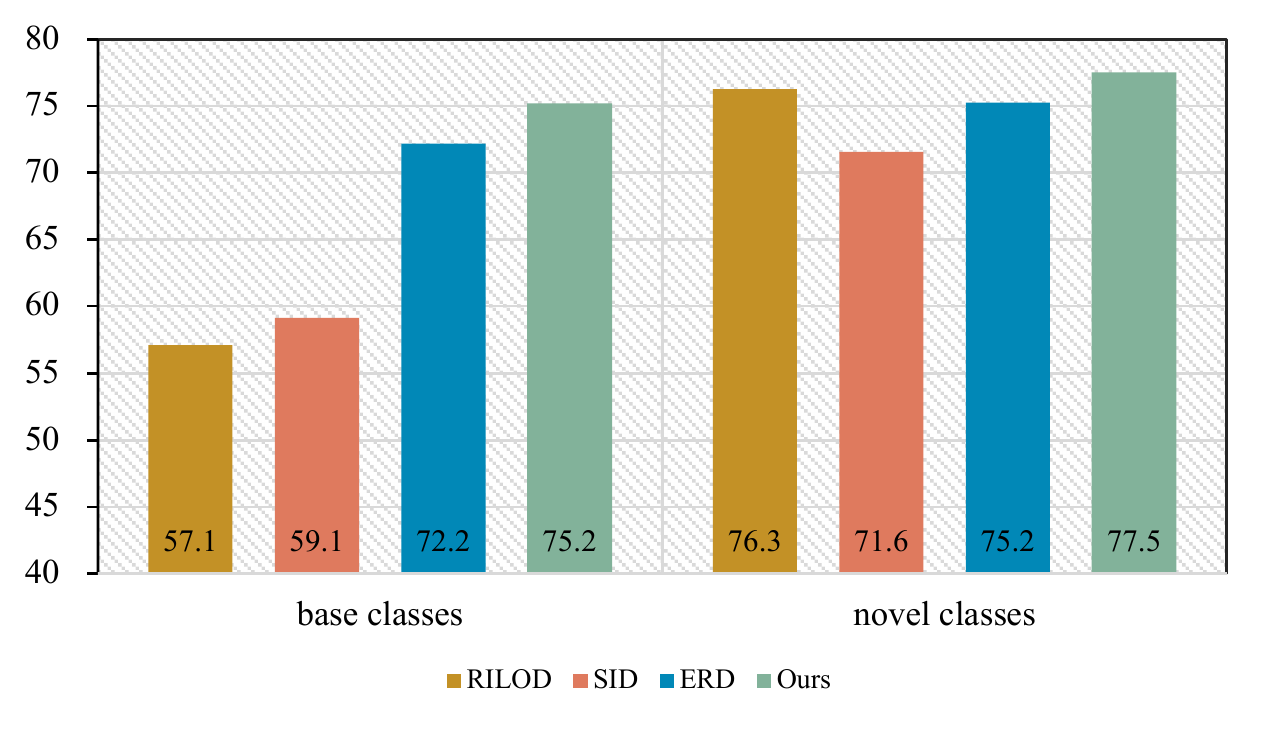}%
\label{fig_base_and_novel_1}}
\\\vspace{-10pt}
\subfloat[\small Average Precision in Step 3]{\includegraphics[width=0.45\textwidth]{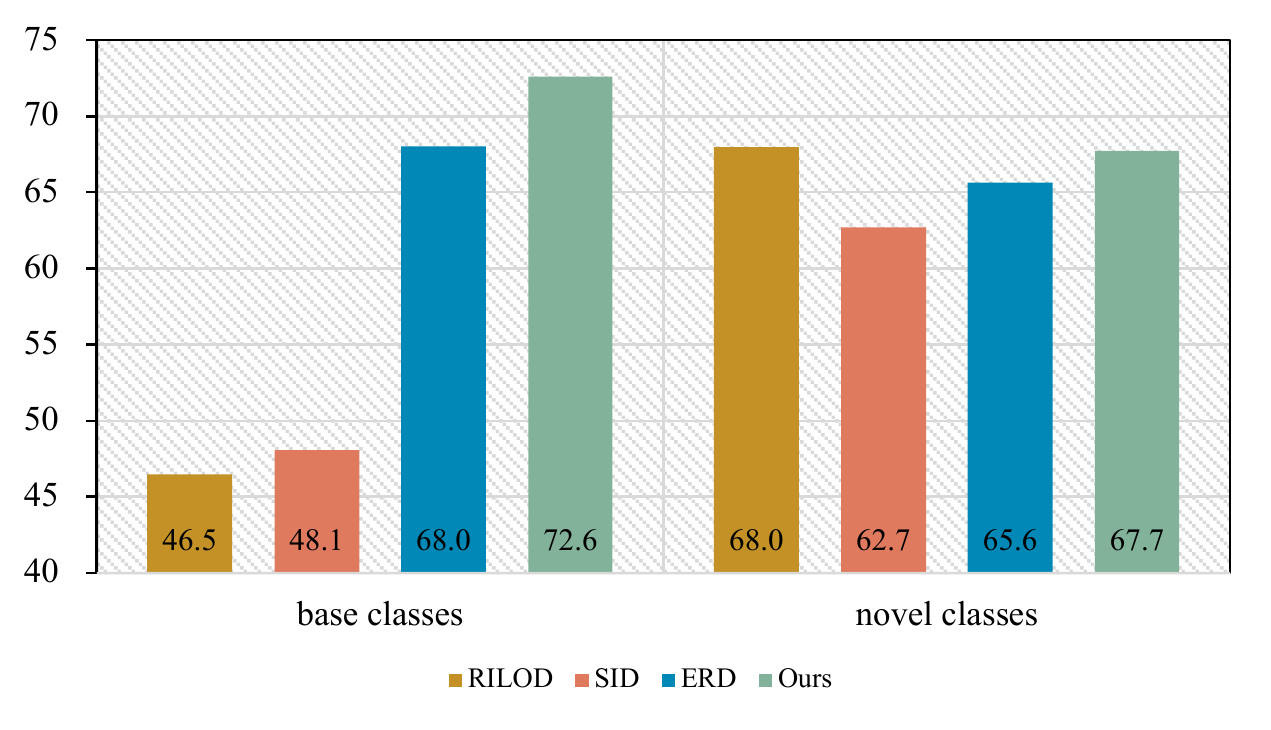}%
\label{fig_base_and_novel_2}}
\caption{\small Results(\%) of base classes and novel classes in each incremental step of the three-step setting on the NBA-IOD benchmark.}
\label {fig_base_and_novel}\vspace{-12pt}
\end{figure}

Figure~\ref{fig_base_and_novel} depicts the average accuracy of base and novel classes across different incremental steps. 
For the base classes, the proposed method attains a higher detection accuracy than RILOD, SID, and ERD in both step 2 (Figure~\ref{fig_base_and_novel_1}) and step 3 (Figure~\ref{fig_base_and_novel_2}).
This outcome suggests that $R^2D$ exhibits a superior ability to preserve the knowledge learned from base classes in incremental player detection.
Concerning the novel classes, $R^2D$ achieves the highest accuracy in step 2 (Figure~\ref{fig_base_and_novel_1}) and attains an accuracy marginally lower than RILOD while surpassing SID and ERD in step 3 (Figure~\ref{fig_base_and_novel_2}).
Although RILOD performs well on novel classes, its accuracy for base classes is underwhelming, resulting in a considerably lower overall performance compared to $R^2D$.
Compared to ERD, our method demonstrates exceptional performance for both base and novel classes throughout all incremental steps in Figures~\ref{fig_base_and_novel_1} and \ref{fig_base_and_novel_2}.
In summary, these results confirm that $R^2D$ effectively maintains learned knowledge from base classes while facilitating the learning of new classes.

\begin{table*}[t]
\centering
\caption{Incremental results(\%) on NBA-IOD benchmark under three-step setting.}
\label{table_three}
\begin{tabular}{@{}l|bbb|ddd|ggg}
\toprule
                         & \multicolumn{3}{b|}{Step 1}                                           & \multicolumn{3}{d|}{Step 2}                   & \multicolumn{3}{g}{Step 3}                    \\ 
\multirow{-2}{*}{Methods}&$AP$                    & $AP_{50}$                  & $AP_{75}$                  &$AP$            & $AP_{50}$          & $AP_{75}$          &$AP$            & $AP_{50}$          & $AP_{75}$ \\ \midrule
Upper Bound              &                       &                      &                        & 77.7          & 93.5          & 89.4          & 78.1          & 94.0          & 90.0 \\
Catastrophic Forgetting  &                       &                       &                       & 41.9          & 50.5          & 48.3          & 38.3          & 46.9          & 44.9 \\
RILOD \cite{refx13}        &                       &                       &                       & 67.0          & 82.6          & 78.4          & 53.8          & 66.4          & 63.0 \\
SID \cite{refx14}         &                       &                       &                       & 65.5          & 82.4          & 77.2          & 53.0          & 66.8          & 62.7 \\
ERD \cite{refx17}          &                       &                       &                       & 73.8          & 90.9          & 86.0          & 67.2          & 82.4          & 78.2 \\
$R^2D$                   &\multirow{-6}{*}{76.0} &\multirow{-6}{*}{96.1} &\multirow{-6}{*}{89.5} & \textbf{76.4} & \textbf{93.4} & \textbf{88.6} & \textbf{70.9} & \textbf{86.2} & \textbf{82.0} \\ 
\bottomrule
\end{tabular}
\end{table*}

{\bf{Five-step setting.}} 
We further present incremental results of the proposed method under the five-step setting. 
Table~\ref{table_five} demonstrates that $R^2D$ exhibits a significant performance improvement across all incremental steps when compared to directly fine-tuning. 
This finding substantiates that our method can considerably mitigate catastrophic forgetting in player IOD tasks, corroborating the previous conclusion from the three-step setting.
Similarly, our method surpasses RILOD and SID, substantially enhancing performance in all incremental steps.
Furthermore, in comparison to ERD, our method increases the $AP$ by 1.8\%, 1.9\%, 2.3\%, and 1.1\%, increases the $AP_{50}$ by 1.4\%, 1.9\%, 2.3\%, and 1.0\%, and increases the $AP_{75}$ by 1.5\%, 2.3\%, 2.4\%, and 1.0\% in each incremental step.
These results indicate that $R^2D$ possesses superior continuous learning capabilities than other methods in incremental player detection.
This is attributed to the proposed refined response distillation strategy, which effectively captures knowledge from different regions, thereby alleviating catastrophic forgetting during incremental learning.

\begin{table*}[t]
\centering
\caption{Incremental results(\%) on NBA-IOD benchmark under five-step setting.}
\label{table_five}
 \resizebox{0.95\textwidth}{!}{
\begin{tabular}{@{}l|bbb|ddd|ggg|eee|fff}
\toprule
                     & \multicolumn{3}{b|}{Step   1}                                         & \multicolumn{3}{d|}{Step   2} & \multicolumn{3}{g|}{Step   3} & \multicolumn{3}{e|}{Step   4} & \multicolumn{3}{f}{Step   5} \\
\multirow{-2}{*}{Methods}&$AP$                    & $AP_{50}$                  & $AP_{75}$                  &$AP$       & $AP_{50}$     & $AP_{75}$    &$AP$       & $AP_{50}$     & $AP_{75}$    &$AP$       & $AP_{50}$     & $AP_{75}$    &$AP$       & $AP_{50}$    & $AP_{75}$    \\ \midrule
Catastrophic Forgetting&                       &                       &                         &31.1 	    &37.7 	   &35.9 	 &21.5 	    &26.2 	   &24.8 	 &25.6 	    &31.4 	   &29.8 	&16.1 	&20.1 	&18.8    \\
RILOD \cite{refx13}      &                       &                       &                       & 69.1     & 85.3     & 81.0    & 67.1     & 81.6     & 77.8    & 50.5     & 61.6     & 58.7    & 51.6     & 62.7    & 60.2    \\
SID \cite{refx14}       &                       &                       &                       & 67.4     & 84.5     & 79.8    & 65.8     & 81.4     & 77.0    & 52.2     & 64.8     & 61.3    & 53.9     & 66.7    & 63.2    \\
ERD \cite{refx17}        &                       &                       &                       & 73.3     & 90.0     & 85.9    & 71.5     & 86.6     & 82.4    & 66.0     & 80.0     & 76.5    & 65.1     & 78.7    & 75.4    \\
$R^2D$                 &\multirow{-5}{*}{76.0} &\multirow{-5}{*}{96.1} & \multirow{-5}{*}{89.5}                      &\textbf{75.1}&\textbf{91.4}&\textbf{87.4}&\textbf{73.4}&\textbf{88.4}&\textbf{84.7}&\textbf{68.3}&\textbf{82.3}&\textbf{78.9}&\textbf{66.2}&\textbf{79.7}&\textbf{76.4}\\ \bottomrule
\end{tabular}}
\end{table*}

{\bf{Home-and-away setting.}} 
In practice, the home-and-away setting is a more prevalent incremental scenario due to the distinct uniforms accommodating different playing fields.
Table~\ref{table_home_and_away} presents the incremental results for this scenario. 
Compared to ERD, the proposed method achieves improvements of 1.3\%, 1.6\%, and 0.5\% on $AP$, $AP_{50}$, and $AP_{75}$, respectively. 
Moreover, our method significantly outperforms directly fine-tuning, SID, and RILOD.
The conclusion is consistent with the one in the multi-step setting.
These results demonstrate the robustness and applicability of $R^2D$ in the home-and-away setting.

\begin{table}[h]
 \centering
 \vspace{-12pt}
\caption{Incremental results (\%) on NBA-IOD benchmark under home-and-away setting.}
\vspace{-6pt}
\label{table_home_and_away}
 \resizebox{0.49\textwidth}{!}{
\begin{tabular}{@{}l|bbb|ddd}
\toprule
                         & \multicolumn{3}{b|}{Step 1}                                           & \multicolumn{3}{d}{Step 2}                    \\ 
\multirow{-2}{*}{Methods}&$AP$                    & $AP_{50}$                  & $AP_{75}$                  &$AP$            & $AP_{50}$          & $AP_{75}$          \\ \midrule
Catastrophic Forgetting  &                      &                       &                       & 41.1          & 52.0          & 48.7          \\
RILOD\cite{refx13}         &                       &                       &                       & 69.3          & 90.7          & 84.6          \\
SID \cite{refx14}         &                       &                       &                       & 65.4          & 89.0          & 80.6          \\
ERD \cite{refx17}          &                       &                       &                       & 71.0          & 91.6          & 86.0          \\
$R^2D$                   & \multirow{-5}{*}{70.3}&\multirow{-5}{*}{94.3} &\multirow{-5}{*}{88.3} & \textbf{72.3} & \textbf{93.2} & \textbf{86.5} \\ \bottomrule
\end{tabular}}\vspace{-6pt}
\end{table}

{\bf{Two-team setting.}} 
In practice, detecting one team in the first step and the other in the incremental step is a flexible and efficient manner to facilitate the player characteristics analysis for two teams competing with each other.
Table~\ref{table_two_ranks} presents the incremental results of $R^2D$ under the two-team setting.
In this case, our method obtains significant performance gains over directly fine-tuning, SID, and RILOD in the incremental step.
Furthermore, compared to ERD, our method achieves improvements of 0.5\%, 0.3\%, and 1.4\% in $AP$, $AP_{50}$, and $AP_{75}$. 
These results demonstrate the broad application capability of the proposed method.

\begin{table}[h]
\centering
 \vspace{-12pt}
\caption{Incremental results(\%) on NBA-IOD benchmark under two-team setting.}
\vspace{-6pt}
\label{table_two_ranks}
 \resizebox{0.49\textwidth}{!}{
\begin{tabular}{@{}l|bbb|ddd}
\toprule
                         & \multicolumn{3}{b|}{Step 1}                                           & \multicolumn{3}{d}{Step 2}                    \\ 
\multirow{-2}{*}{Methods}&$AP$                    & $AP_{50}$                  & $AP_{75}$                  &$AP$            & $AP_{50}$          & $AP_{75}$          \\ \midrule
Catastrophic Forgetting  &                       &                       &                       & 44.8          & 56.1          & 51.6          \\
RILOD\cite{refx13}         &                       &                       &                       & 76.1          & 94.2          & 89.2          \\
SID\cite{refx14}          &                       &                       &                       & 73.6          & 93.7          & 88.1          \\
ERD \cite{refx17}          &                       &                       &                       & 77.6          & 95.9          & 90.8          \\
$R^2D$                   & \multirow{-5}{*}{74.9}&\multirow{-5}{*}{96.2} &\multirow{-5}{*}{89.6} & \textbf{78.1} & \textbf{96.2} & \textbf{92.2} \\ \bottomrule
\end{tabular}} \vspace{-6pt}
\end{table}

{\bf{Volleyball scenario.}} 
To further demonstrate the efficacy of our approach, we conduct an evaluation on the Volleyball-IOD benchmark.
Table~\ref{table_volleyball} exhibits the incremental results, revealing that our method surpasses RILOD, SID, and ERD in terms of $AP$, $AP_{50}$, and $AP_{75}$ in the incremental step.
These findings highlight the robustness of our proposed technique in incremental player detection across diverse sports categories.

\begin{table}[h]
\centering
\caption{Incremental results(\%) on Volleyball-IOD benchmark.}
\vspace{-6pt}
\label{table_volleyball}
 \resizebox{0.49\textwidth}{!}{
\begin{tabular}{@{}l|bbb|ddd}
\toprule
                         & \multicolumn{3}{b|}{Step 1}                                           & \multicolumn{3}{d}{Step 2}                    \\ 
\multirow{-2}{*}{Methods}&$AP$                    & $AP_{50}$                  & $AP_{75}$                  &$AP$            & $AP_{50}$          & $AP_{75}$          \\ \midrule
Catastrophic Forgetting  &                       &                       &                       & 30.3          & 44.4          & 36.3          \\
RILOD\cite{refx13}         &                       &                       &                       & 50.1          & 72.3          & 61.1          \\
SID\cite{refx14}          &                       &                       &                       & 49.8          & 73.7          & 62.5          \\
ERD \cite{refx17}          &                       &                       &                       & 51.2          & 73.9          & 62.1          \\
$R^2D$                   &\multirow{-5}{*}{49.0} &\multirow{-5}{*}{74.0} & \multirow{-5}{*}{57.7} & \textbf{51.5} & \textbf{75.2} & \textbf{62.7} \\ \bottomrule
\end{tabular}} \vspace{-12pt}
\end{table}

\begin{figure*}[!htb]
\centering
\subfloat[\small Upper Bound]{\includegraphics[width=0.40\textwidth]{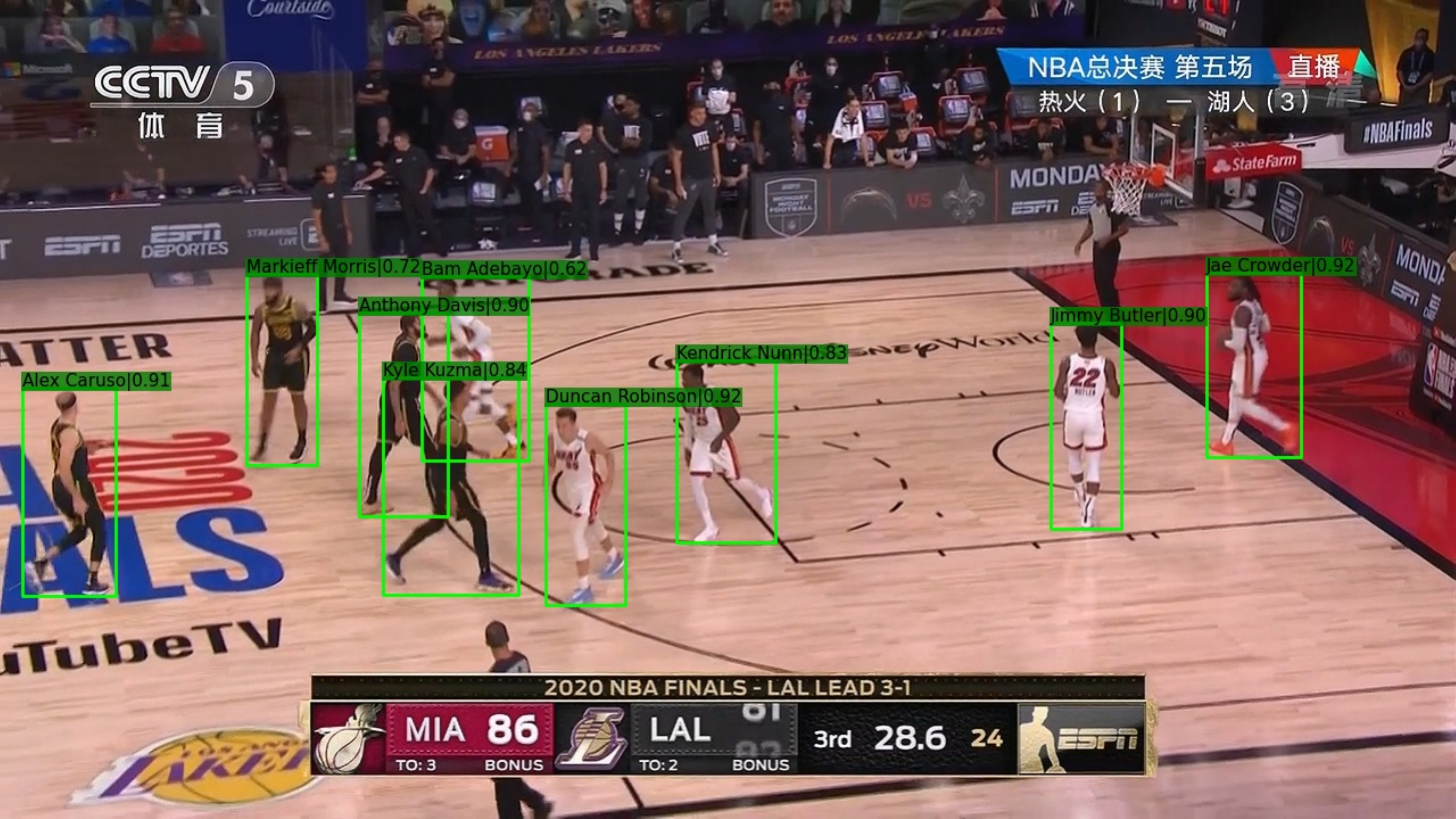}%
\label{fig_three_1}}
\hfil
\subfloat[\small Fine-tuning]{\includegraphics[width=0.40\textwidth]{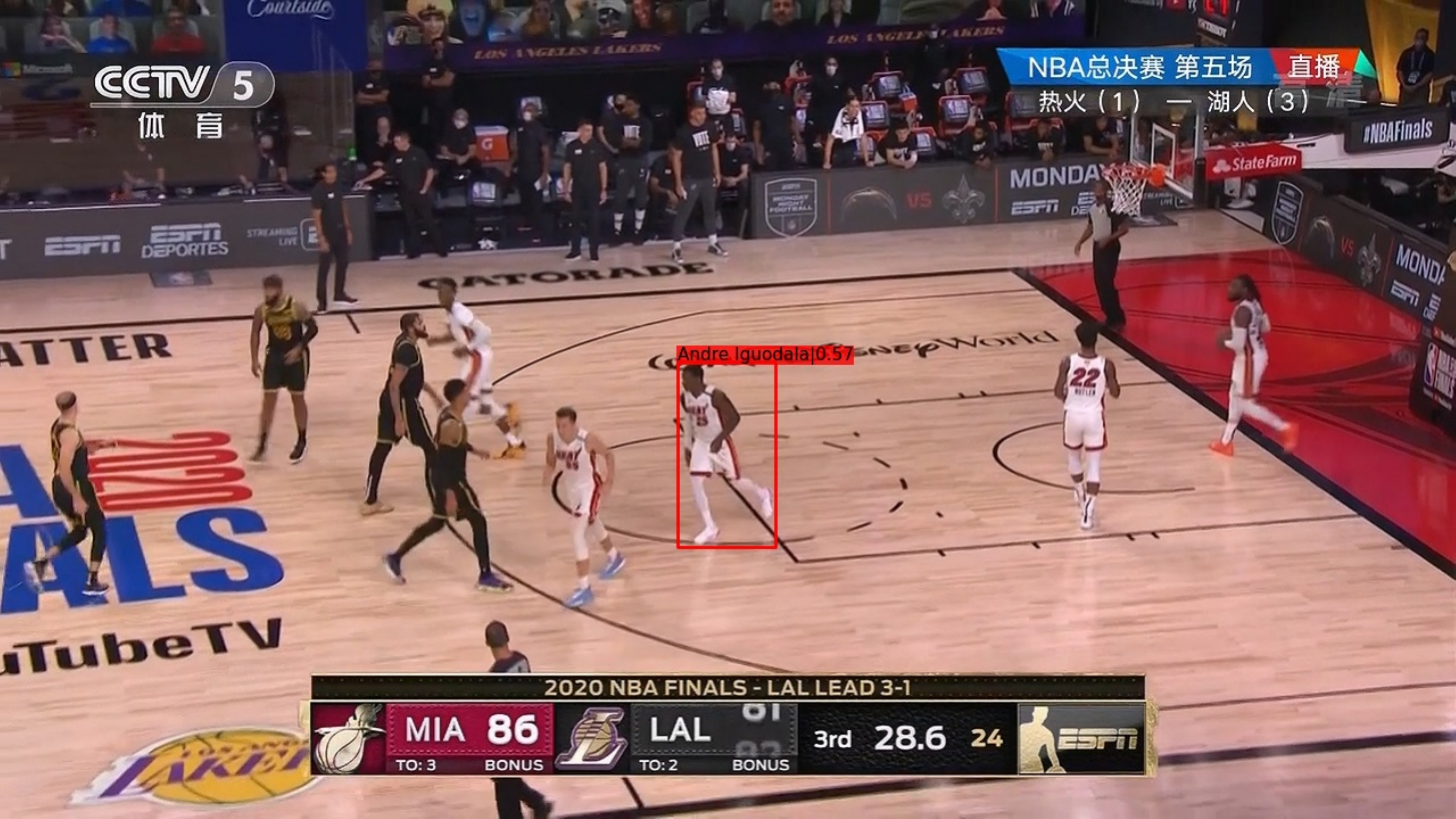}%
\label{fig_three_2}}\\\vspace{-1em}
\subfloat[\small ERD]{\includegraphics[width=0.40\textwidth]{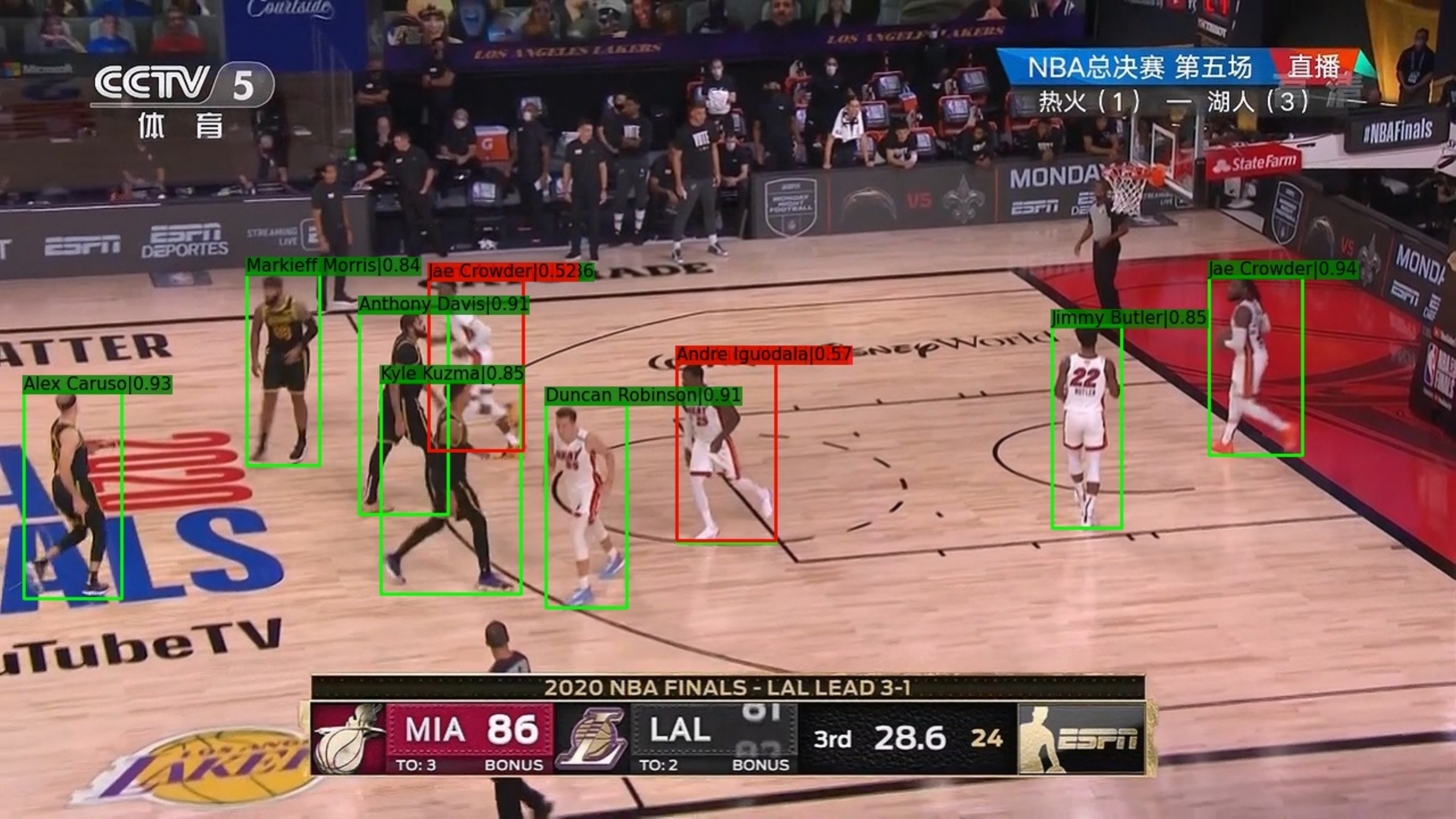}%
\label{fig_three_3}}
\hfil
\subfloat[\small $R^2D$]{\includegraphics[width=0.40\textwidth]{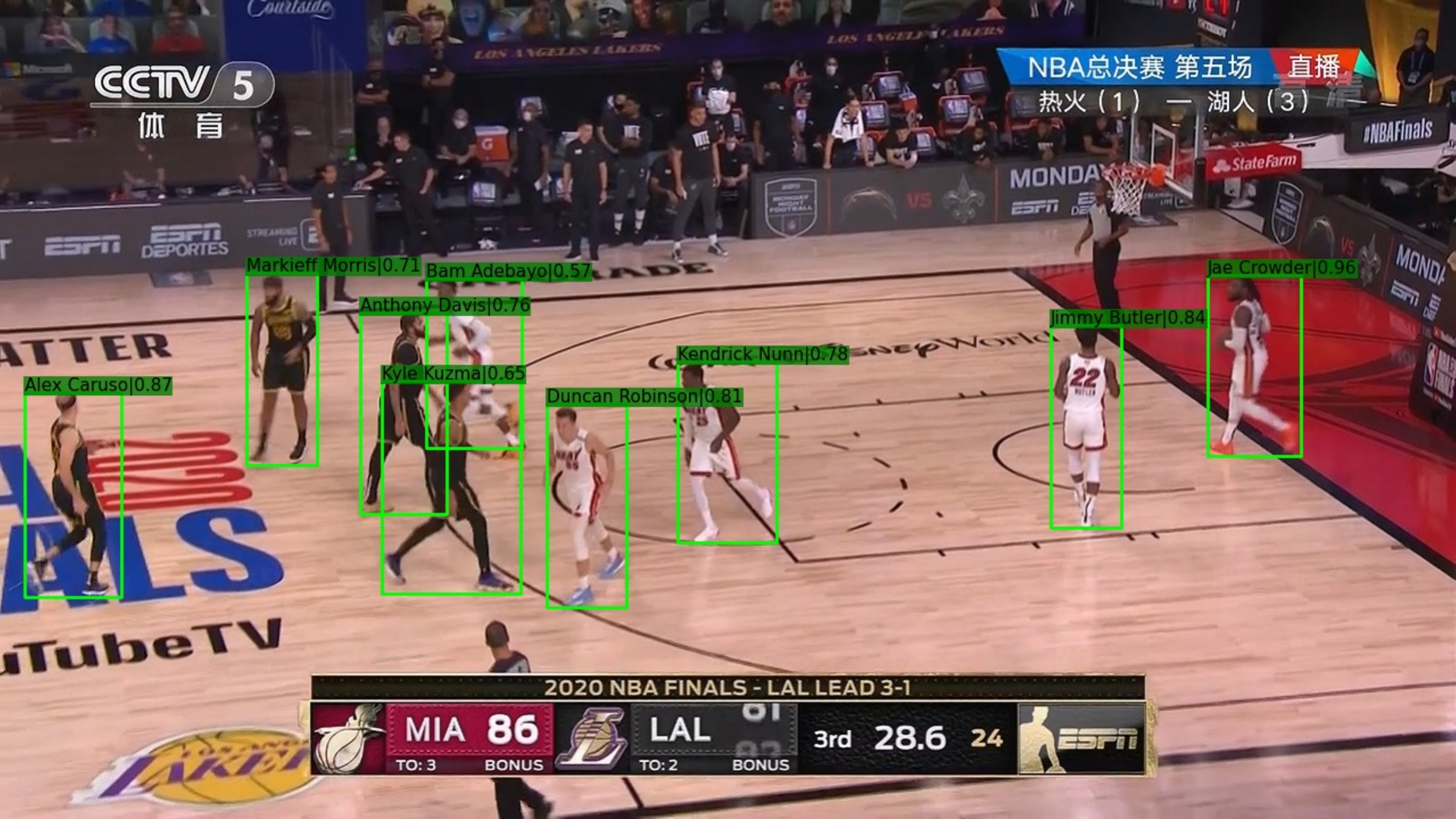}%
\label{fig_three_4}}\vspace{-0.5em}
\caption{\small Visualization of experimental results on NBA-IOD benchmark under the three-step setting: (a) Detection results with GFL training on all data; (b) Detection results with GFL by directly fine-tuning  (Catastrophic Forgetting); (c) Incremental results of ERD; (d) Incremental results of $R^2D$.}\vspace{-12pt}
\label{fig_three}
\end{figure*}
\begin{figure*}[!htb]
\centering
\subfloat[\small Upper Bound]{\includegraphics[width=0.40\textwidth]{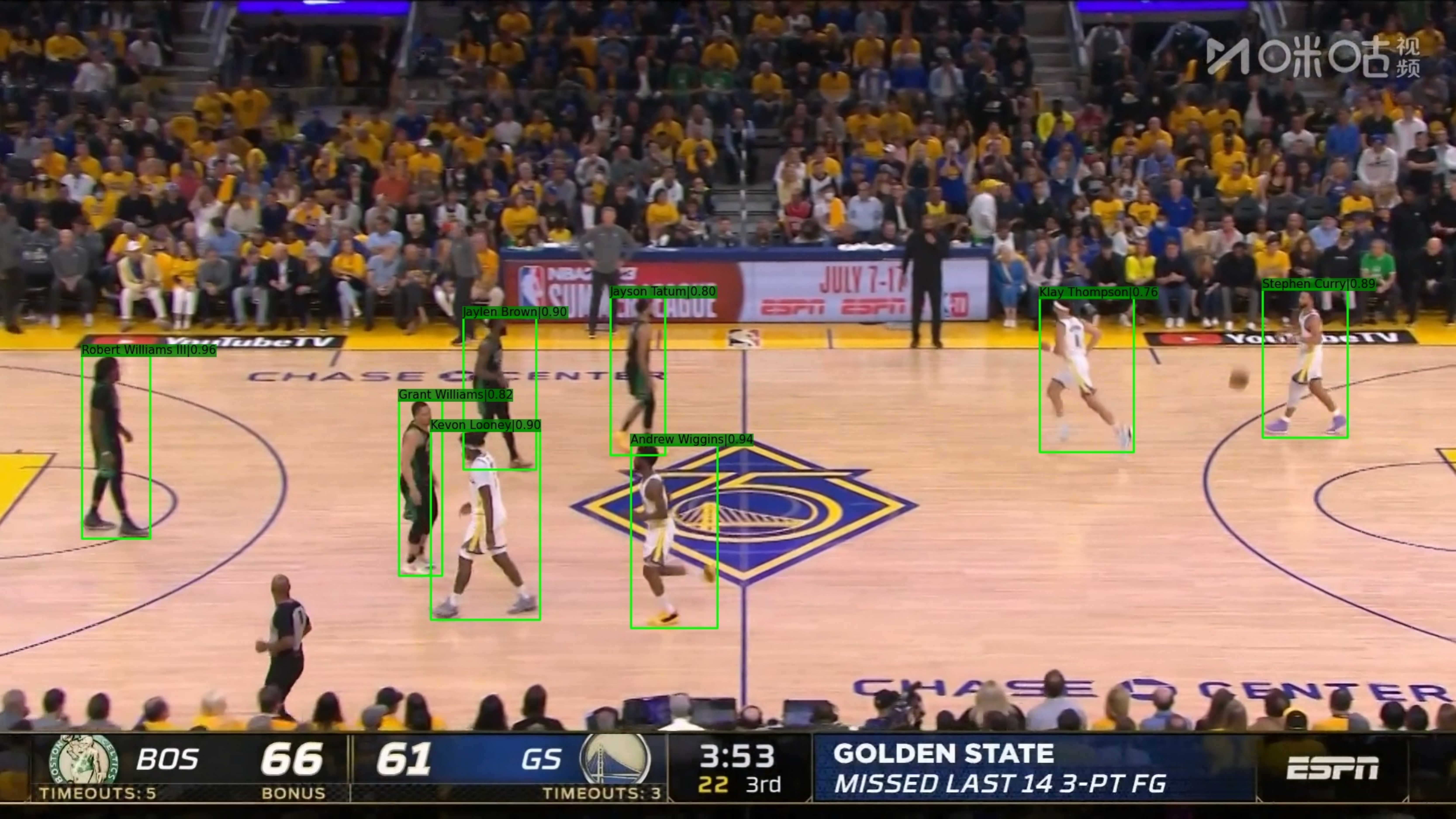}%
\label{fig_five_1}}
\hfil
\subfloat[\small Fine-tuning]{\includegraphics[width=0.40\textwidth]{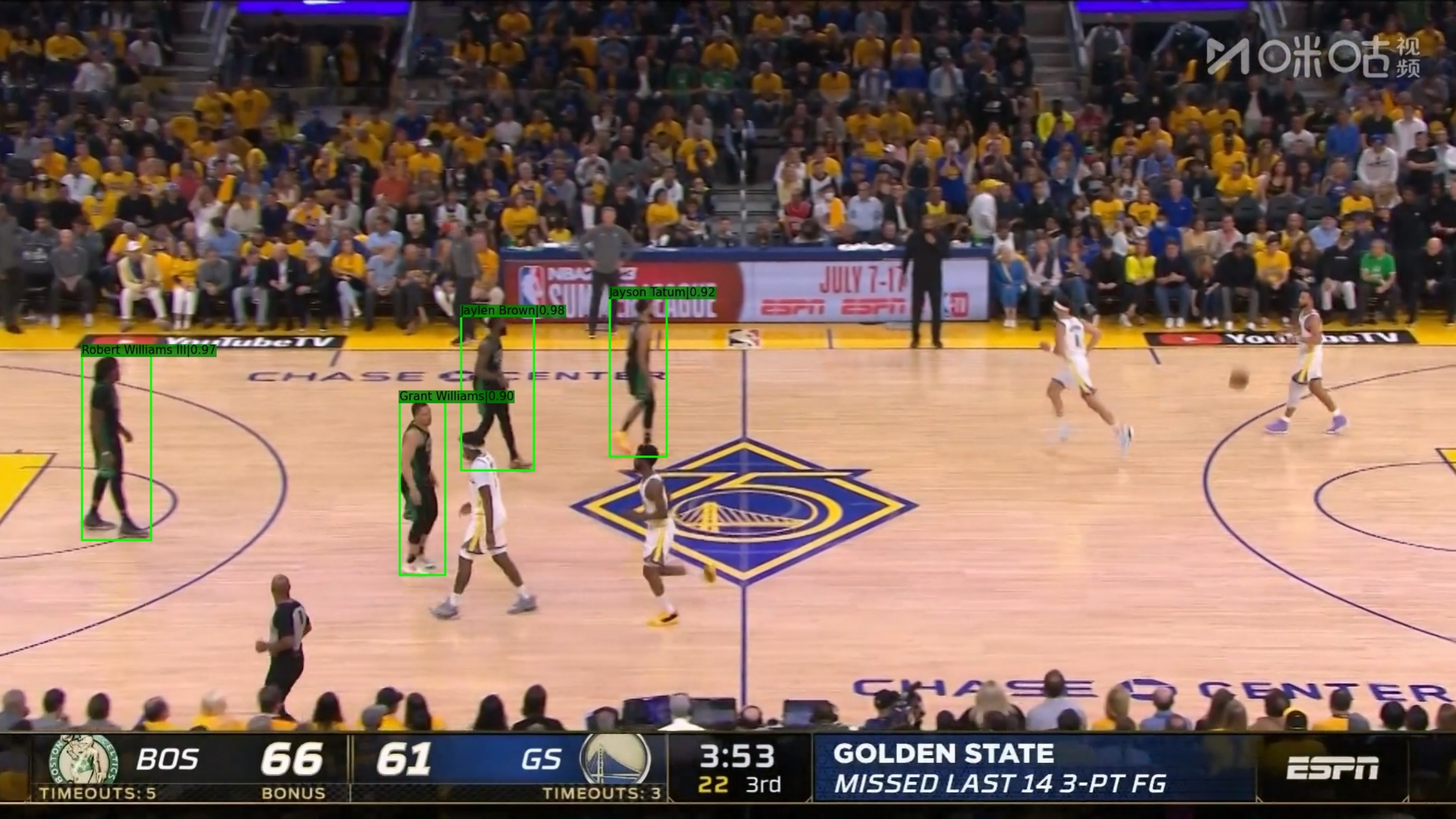}%
\label{fig_five_2}}\\\vspace{-1em}
\subfloat[\small ERD]{\includegraphics[width=0.40\textwidth]{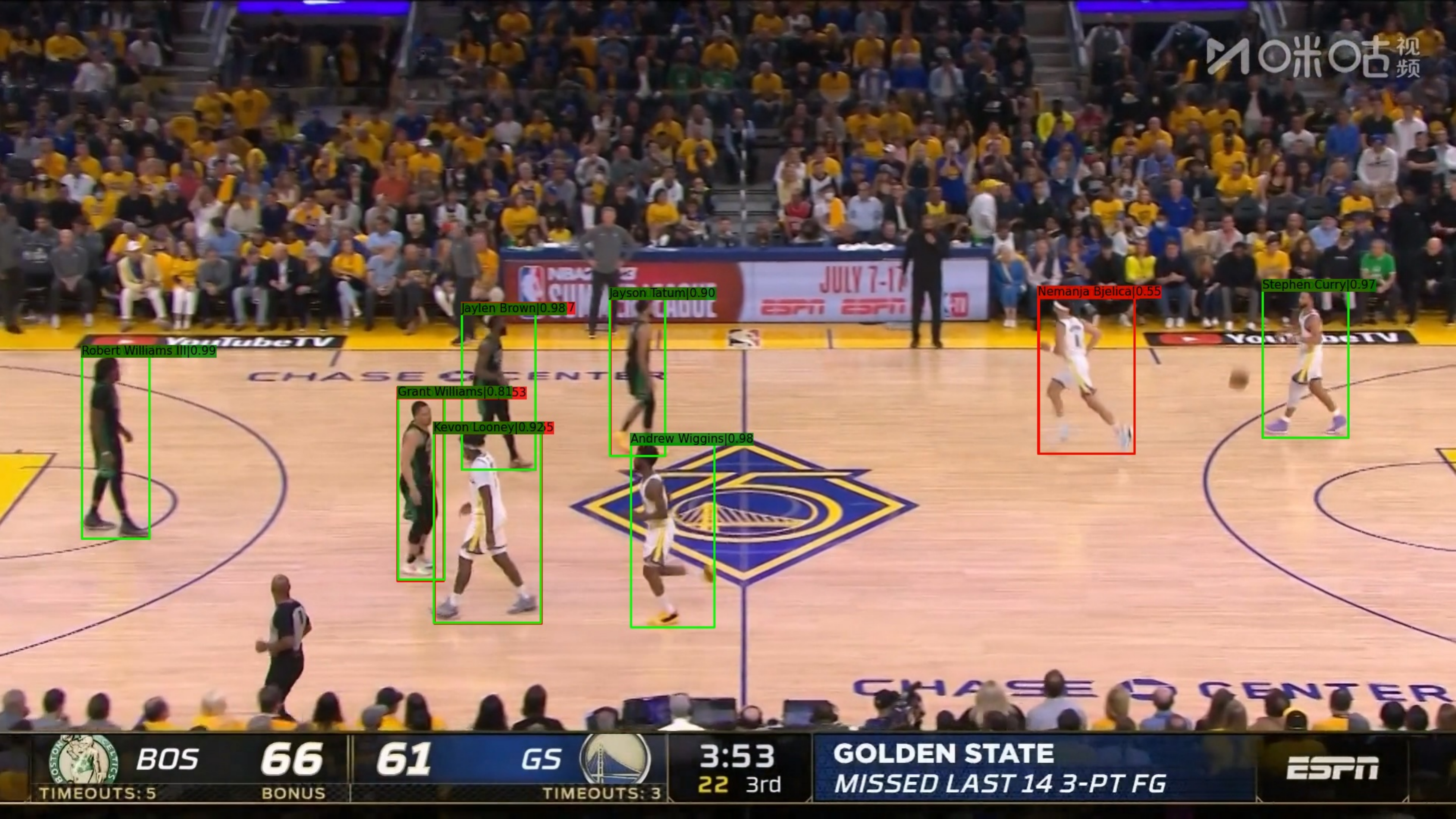}%
\label{fig_five_3}}
\hfil
\subfloat[\small $R^2D$]{\includegraphics[width=0.40\textwidth]{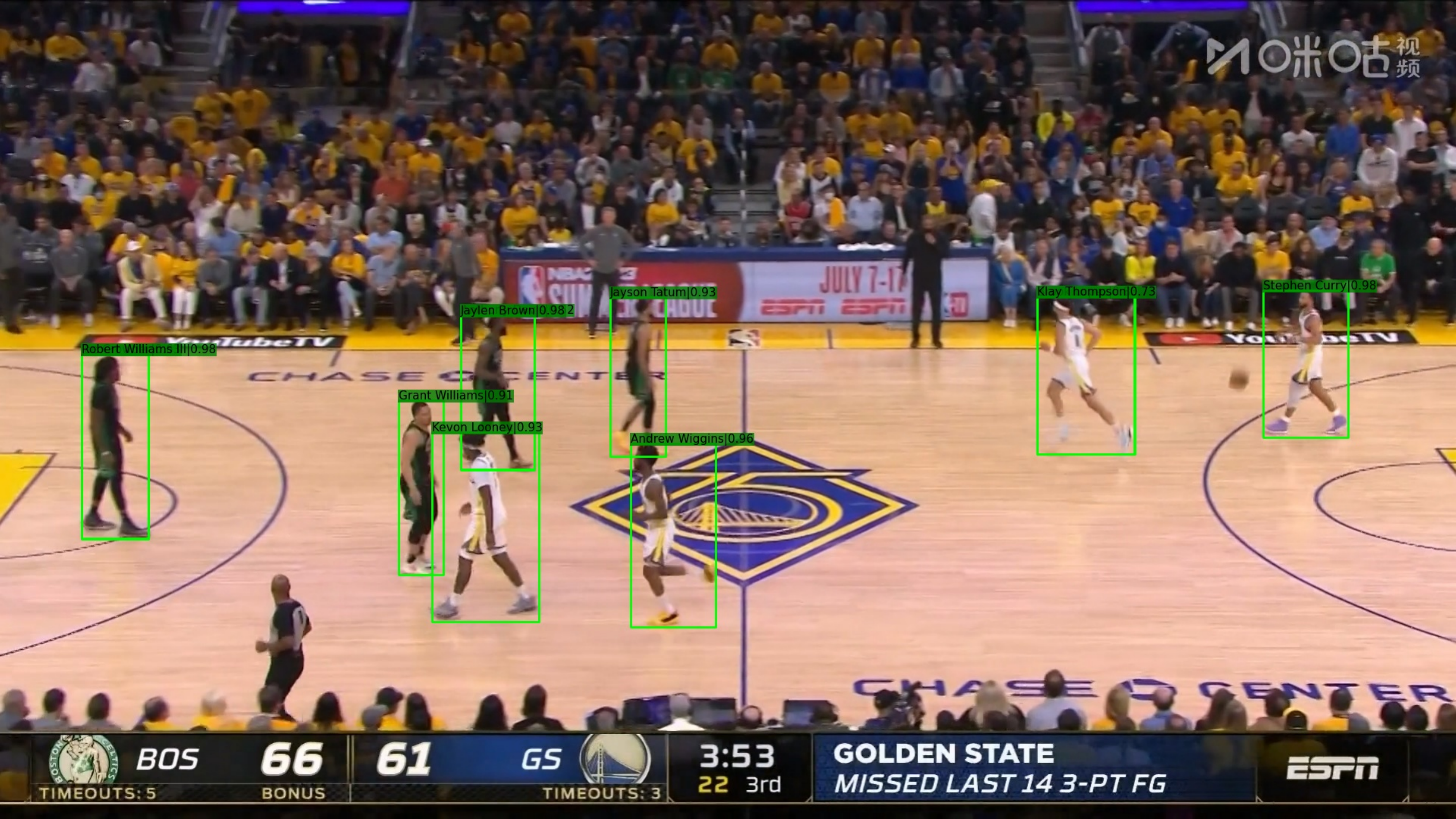}%
\label{fig_five_4}}\vspace{-0.5em}
\caption{\small Visualization of experimental results on NBA-IOD benchmark under the five-step setting: (a) Detection results with GFL training on all data; (b) Detection results with GFL by directly fine-tuning  (Catastrophic Forgetting); (c) Incremental results of ERD; (d) Incremental results of $R^2D$.}\vspace{-12pt}
\label{fig_five}
\end{figure*}

\subsection{Qualitative Results}
Figures \ref{fig_three} and \ref{fig_five} present the visualization results of one sample using different methods under the three-step and five-step settings, respectively. 
Green bounding boxes represent correctly detected players, while red bounding boxes represent incorrectly detected players. 
The text above the boxes shows the detected results with the confidence level.
In Figure~\ref{fig_three_2}, since it has learned the categories of players in the previous but not the latest step, directly fine-tuning barely detects players due to catastrophic forgetting, and the only detection result has an error identity.
In Figure~\ref{fig_five_2}, directly fine-tuning has learned players of the Los Angeles Lakers in the latest step, so only players of this team are detected, and players of the Miami Heat learned previously are forgotten catastrophically.
Notably, as depicted in Figures \ref{fig_three_4} and \ref{fig_five_4}, our proposed method demonstrates favorable performance for all these players and significantly outperforms directly fine-tuning.
As shown in Figures~\ref{fig_three_3} and \ref{fig_five_3},  ERD presents errors and missed detections due to an insufficiently refined incremental distillation strategy, while $R^2D$ performs considerably better (Figures~\ref{fig_three_4} and \ref{fig_five_4}) and performs most closely to the Upper Bound.

\begin{figure*}[!htb]
\centering
\subfloat[\small ERD]{\includegraphics[width=0.40\textwidth]{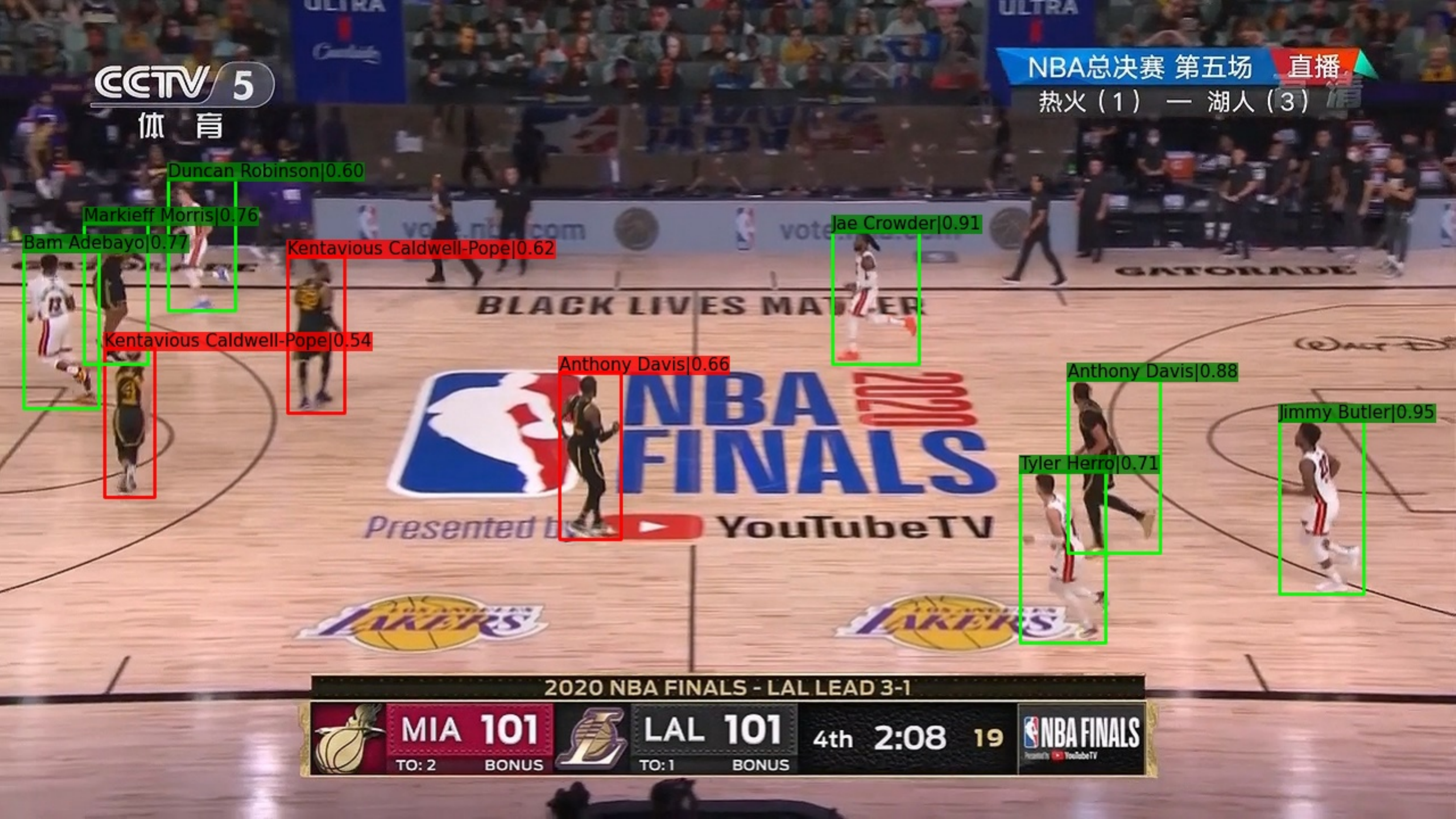}%
\label{fig_home_and_away_1_a}}
\hfil
\subfloat[\small $R^2D$]{\includegraphics[width=0.40\textwidth]{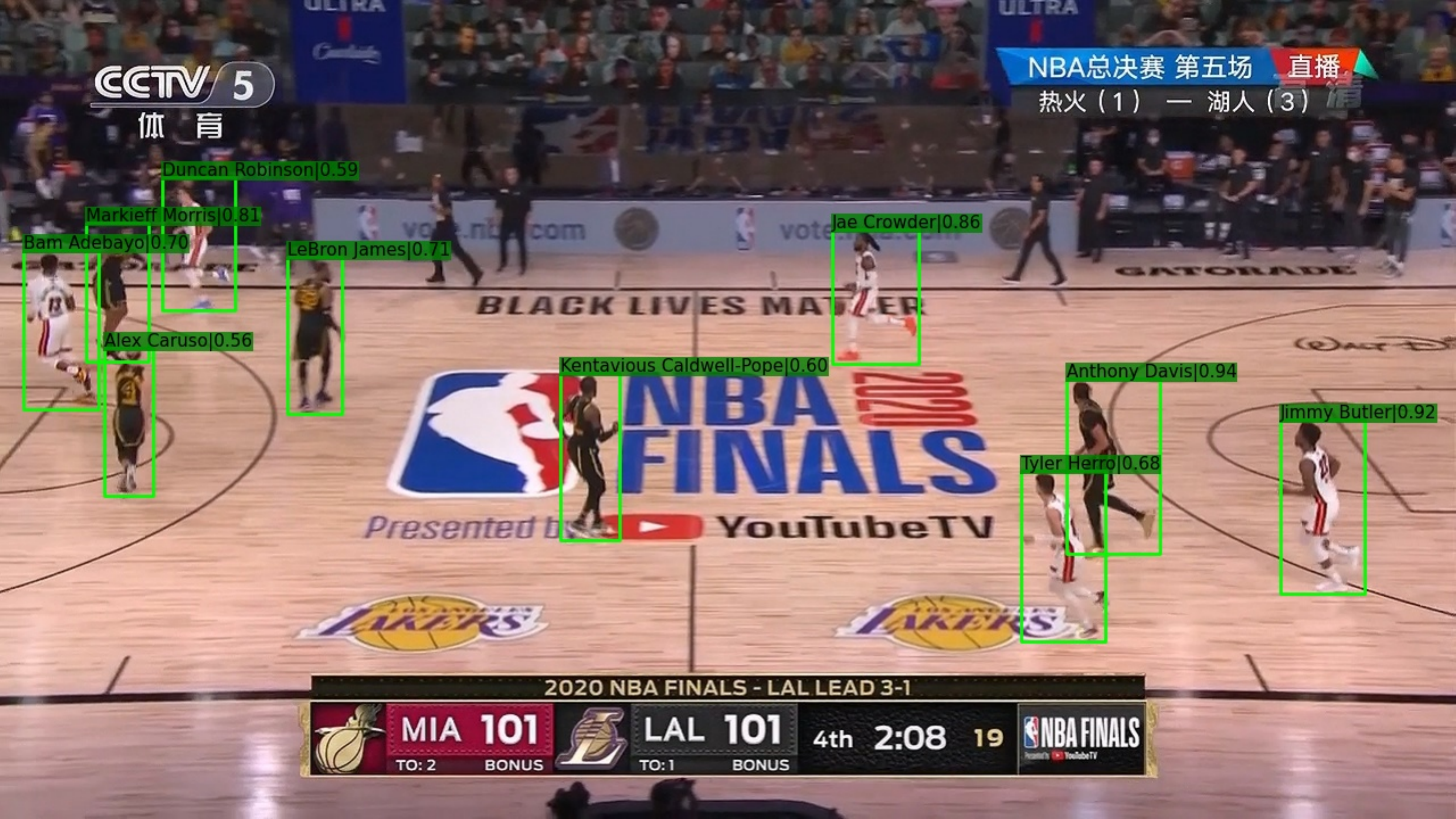}%
\label{fig_home_and_away_1_b}}\\\vspace{-1em}
\subfloat[\small ERD]{\includegraphics[width=0.40\textwidth]{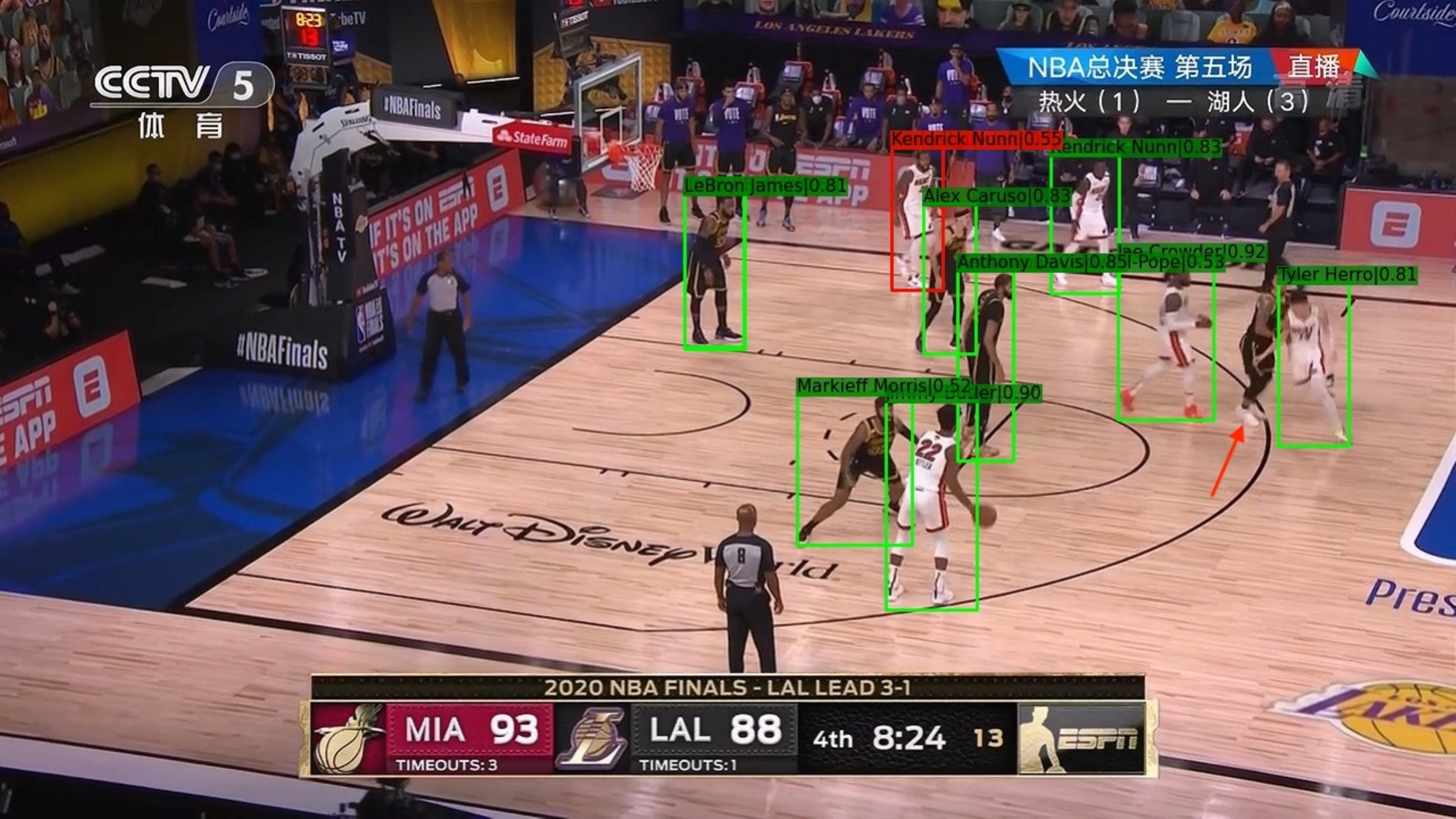}%
\label{fig_home_and_away_2_a}}
\hfil
\subfloat[\small $R^2D$]{\includegraphics[width=0.40\textwidth]{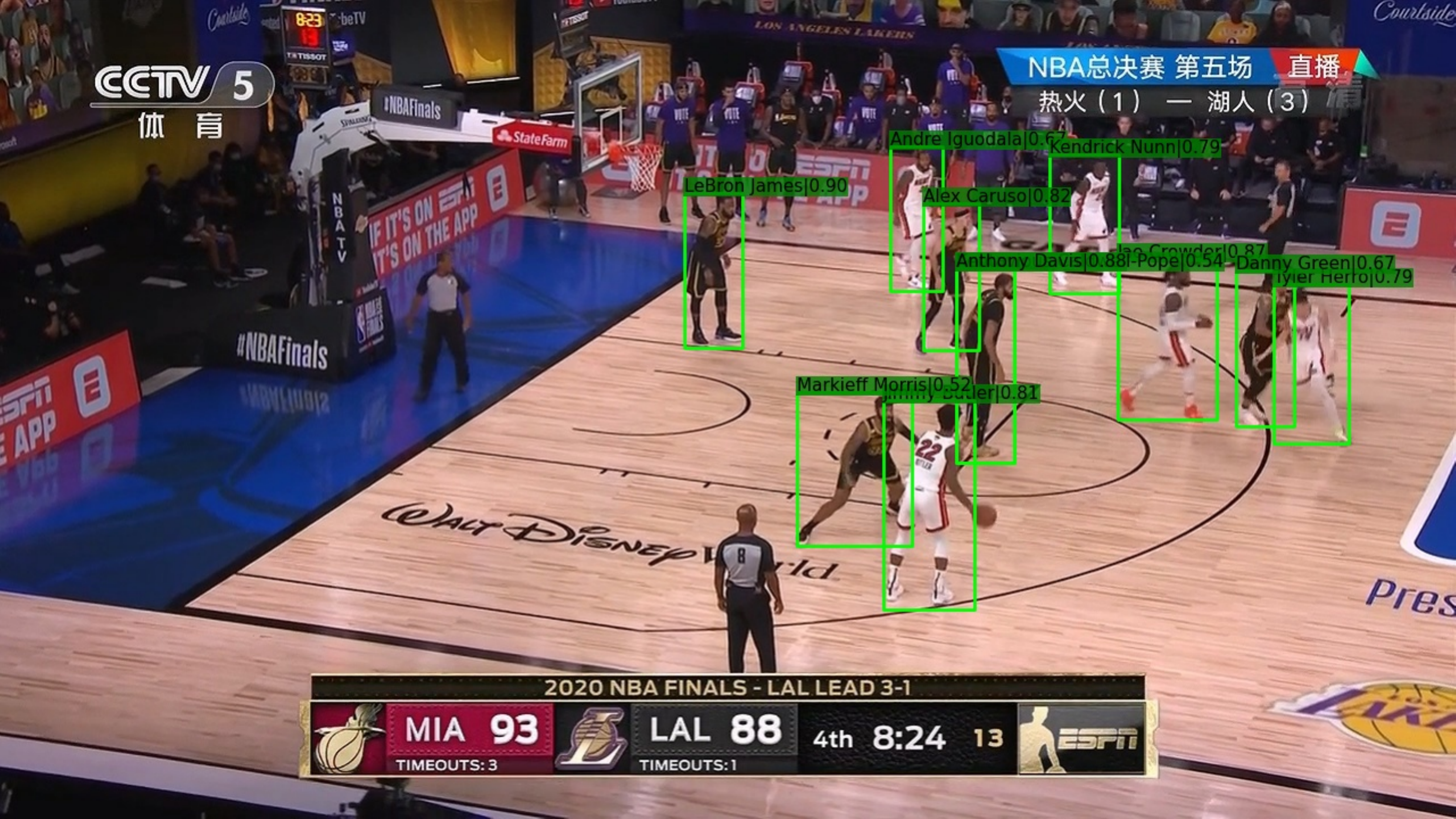}%
\label{fig_home_and_away_2_b}}\vspace{-0.5em}
\caption{\small Visualization of experimental results on NBA-IOD benchmark under home-and-away setting. (a) and (c) are two samples in the incremental results of ERD; (b) and (d) are two samples in the incremental results of $R^2D$.}\vspace{-12pt}
\label{fig_home_and_away}
\end{figure*}

\begin{figure*}[!htb]
\centering
\subfloat[\small ERD]{\includegraphics[width=0.40\textwidth]{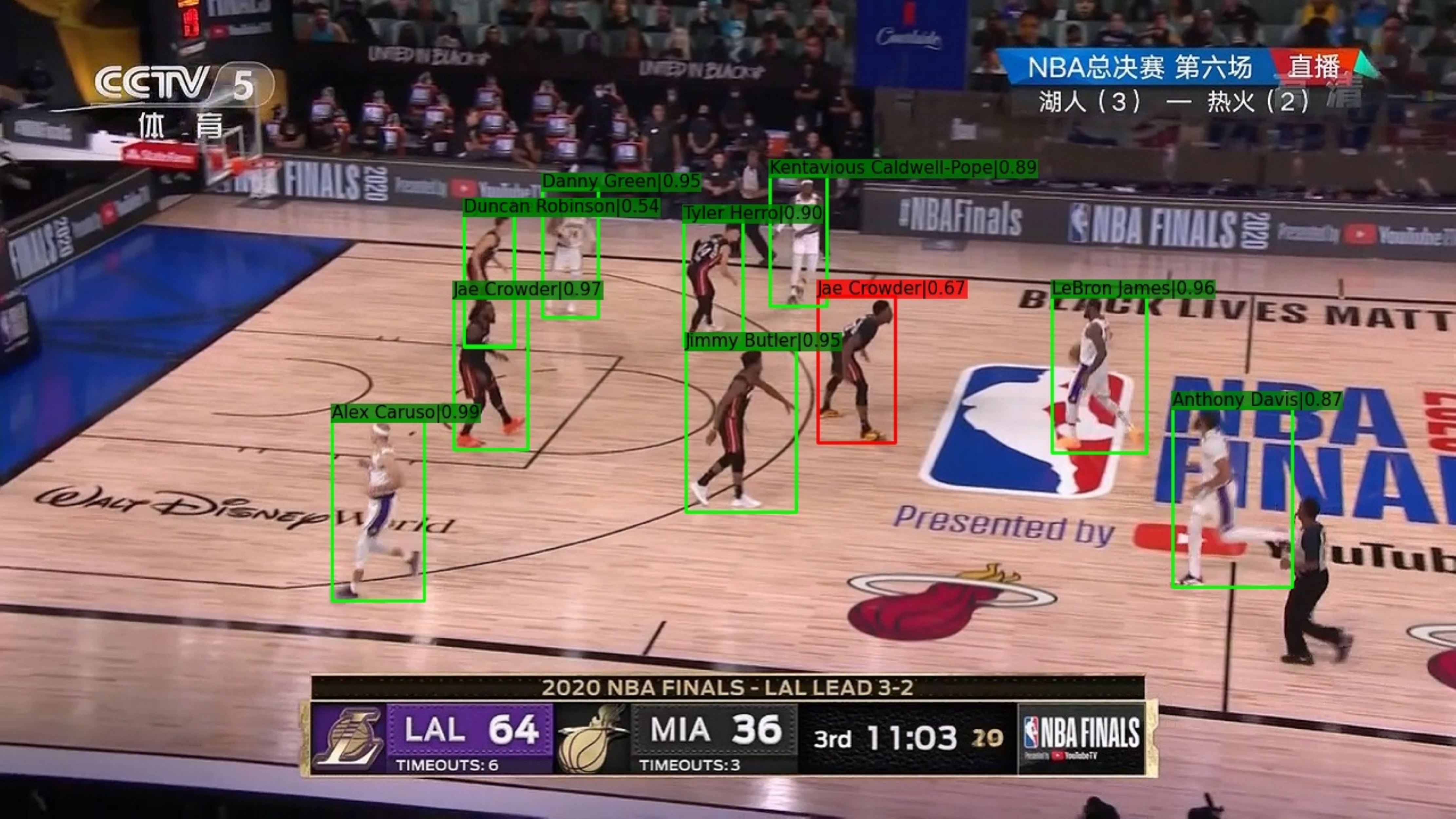}%
\label{fig_two_ranks_1_a}}
\hfil
\subfloat[\small $R^2D$]{\includegraphics[width=0.40\textwidth]{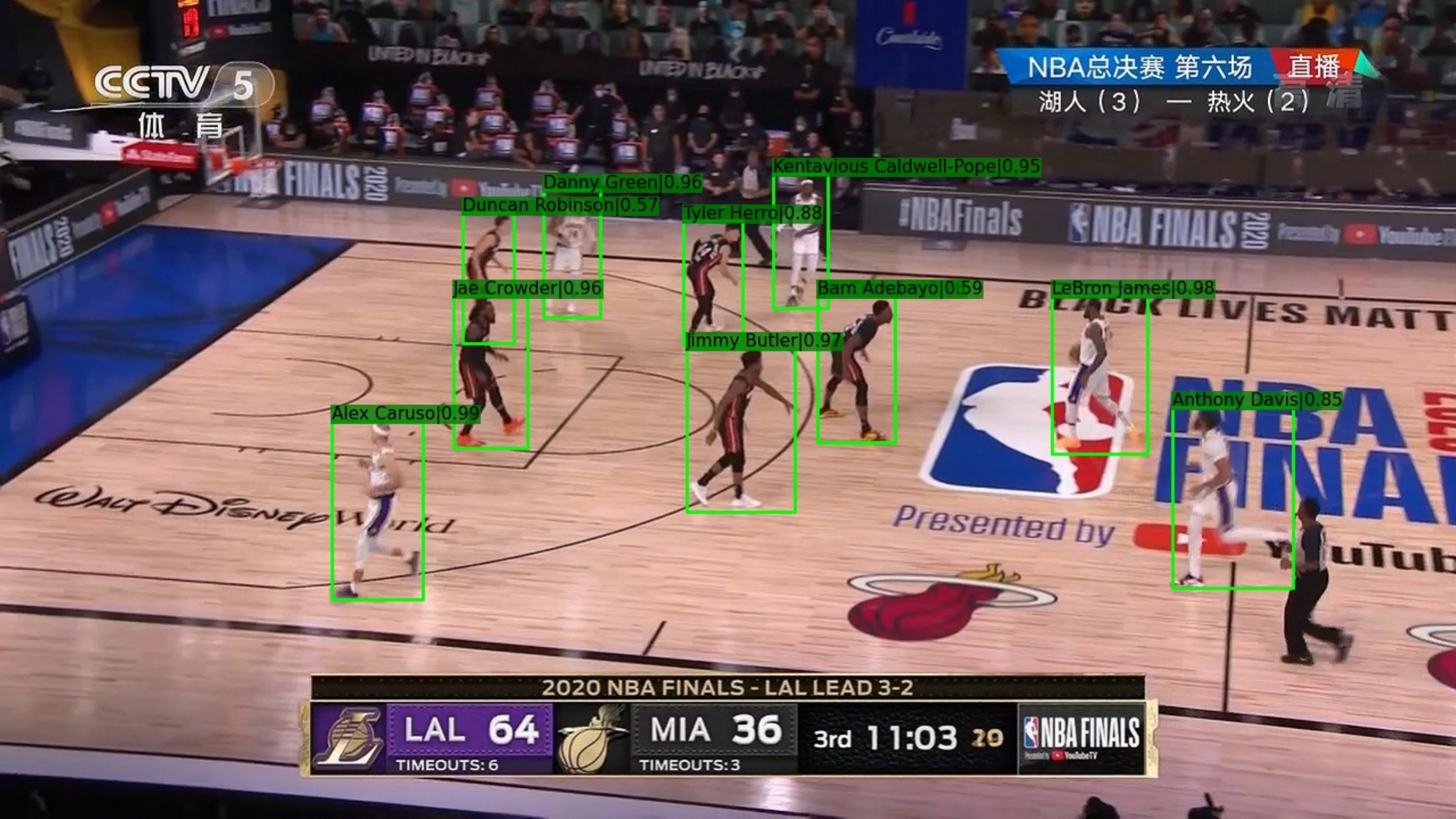}%
\label{fig_two_ranks_1_b}}\\\vspace{-1em}
\subfloat[\small ERD]{\includegraphics[width=0.40\textwidth]{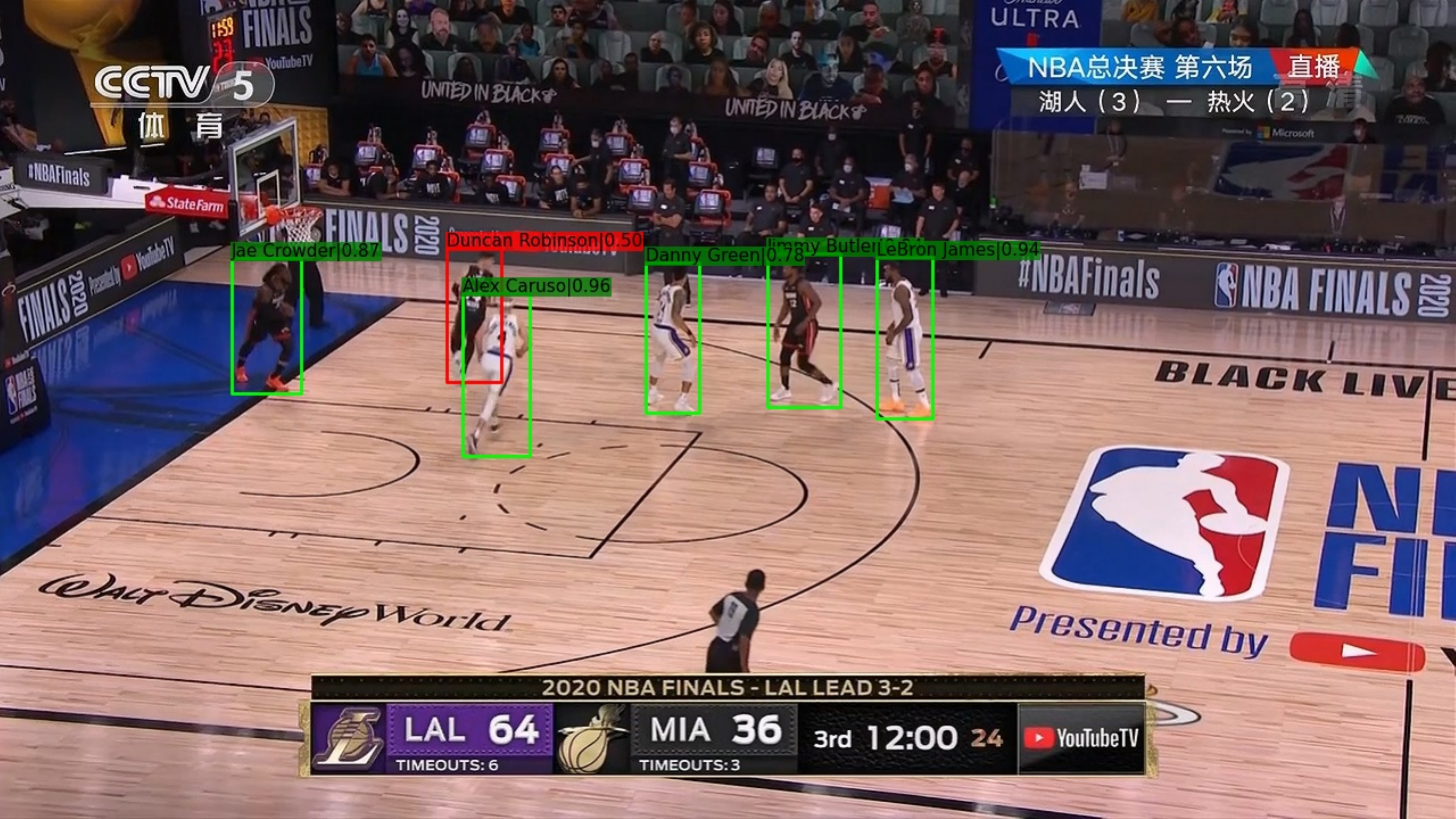}%
\label{fig_two_ranks_2_a}}
\hfil
\subfloat[\small $R^2D$]{\includegraphics[width=0.40\textwidth]{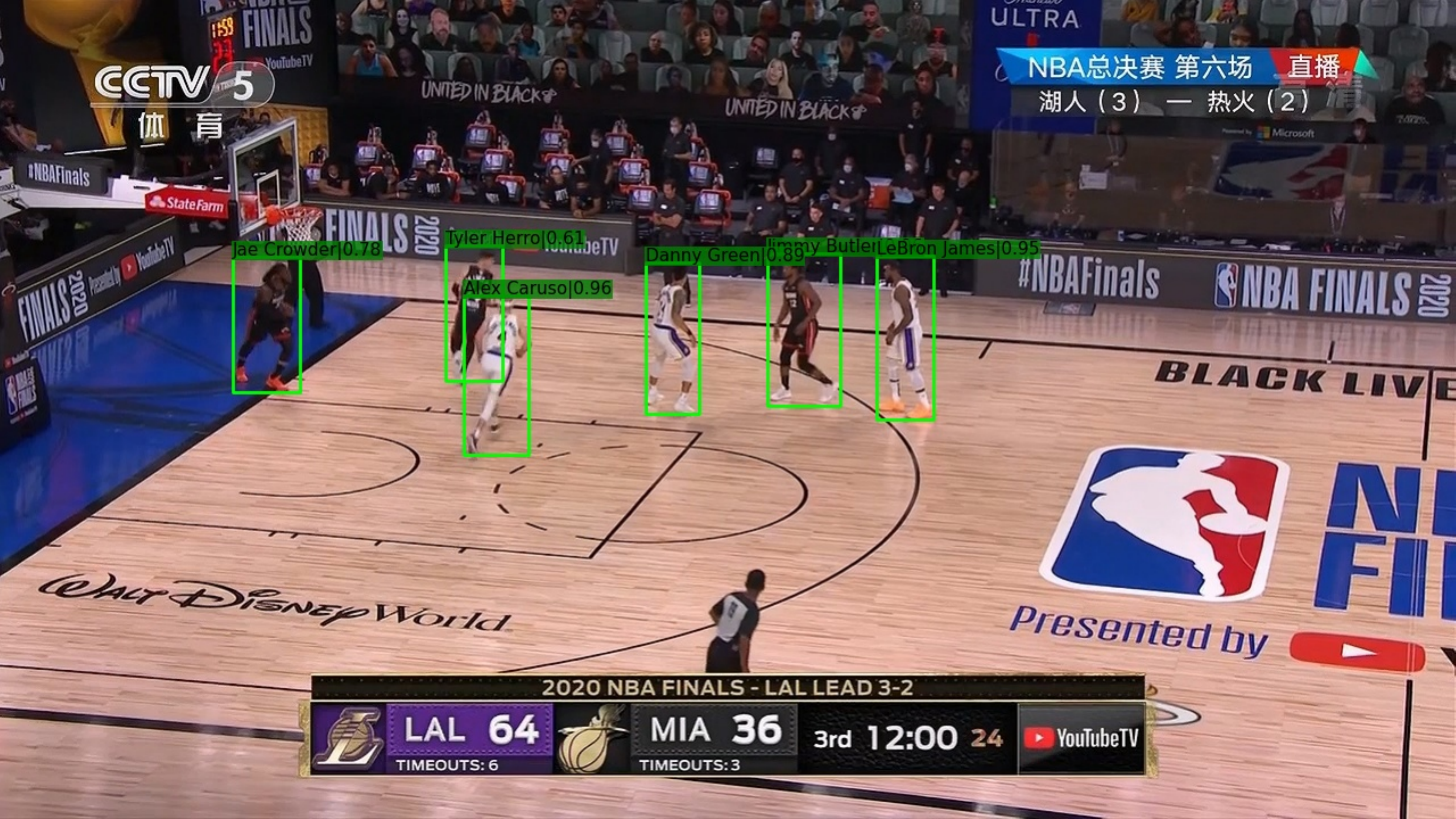}%
\label{fig_two_ranks_2_b}}\vspace{-0.5em}
\caption{\small Visualization of experimental results on NBA-IOD benchmark under two-team setting. (a) and (c) are two samples in the incremental results of ERD; (b) and (d) are two samples in the incremental results of $R^2D$.}\vspace{-12pt}
\label{fig_two_ranks}
\end{figure*}

We further present the visualization results of the proposed method and ERD in the home-and-away setting (Figure~\ref{fig_home_and_away}) and the two-team setting (Figure~\ref{fig_two_ranks}). 
The red arrow indicates that players there are not detected. 
We observe that ERD reports more bad cases although it retains knowledge of the players from previous tasks, $R^2D$ benefits from a refined response distillation strategy that correctly detects the players and efficiently maintains old knowledge. 
In summary, $R^2D$ performs better than the other methods in the IOD tasks of the players, which is consistent with the previous quantitative analysis conclusions.

\begin{figure*}[!htb]
\centering
\subfloat[\small Upper Bound]{\includegraphics[width=0.40\textwidth]{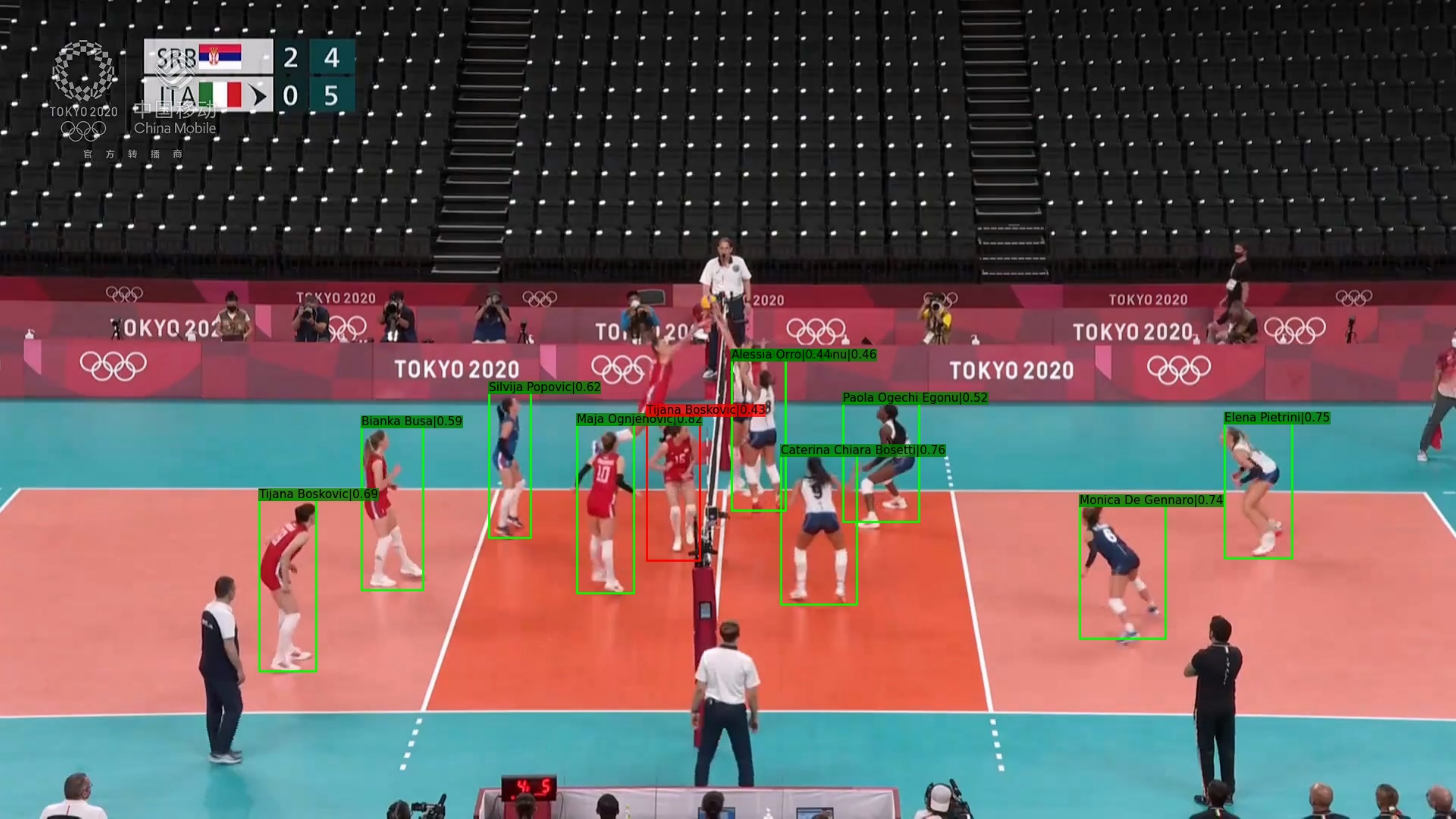}%
\label{fig_volleyball_1}}
\hfil
\subfloat[\small Fine-tuning]{\includegraphics[width=0.40\textwidth]{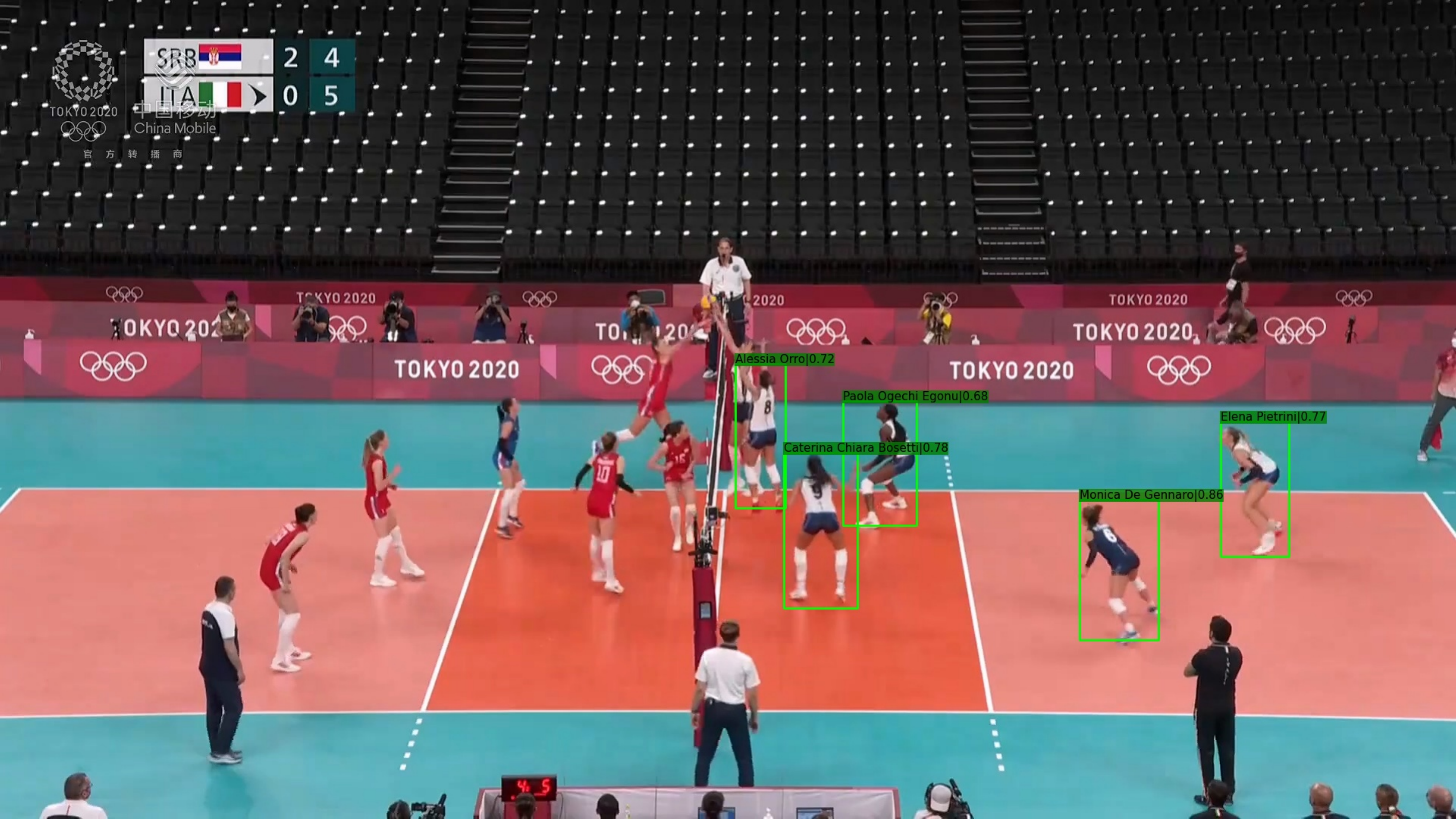}%
\label{fig_volleyball_2}}\\\vspace{-1em}
\subfloat[\small ERD]{\includegraphics[width=0.40\textwidth]{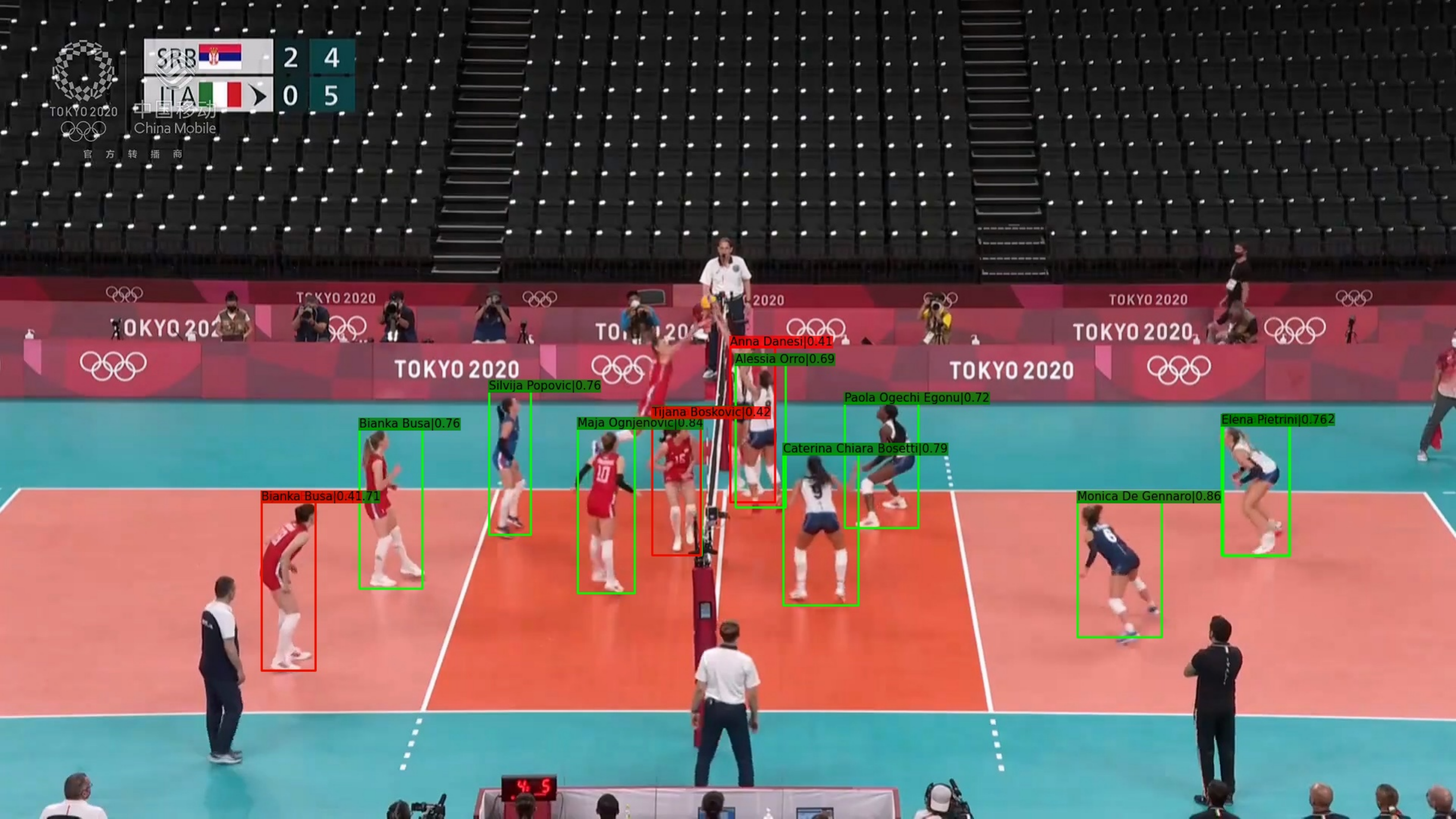}%
\label{fig_volleyball_3}}
\hfil
\subfloat[\small $R^2D$]{\includegraphics[width=0.40\textwidth]{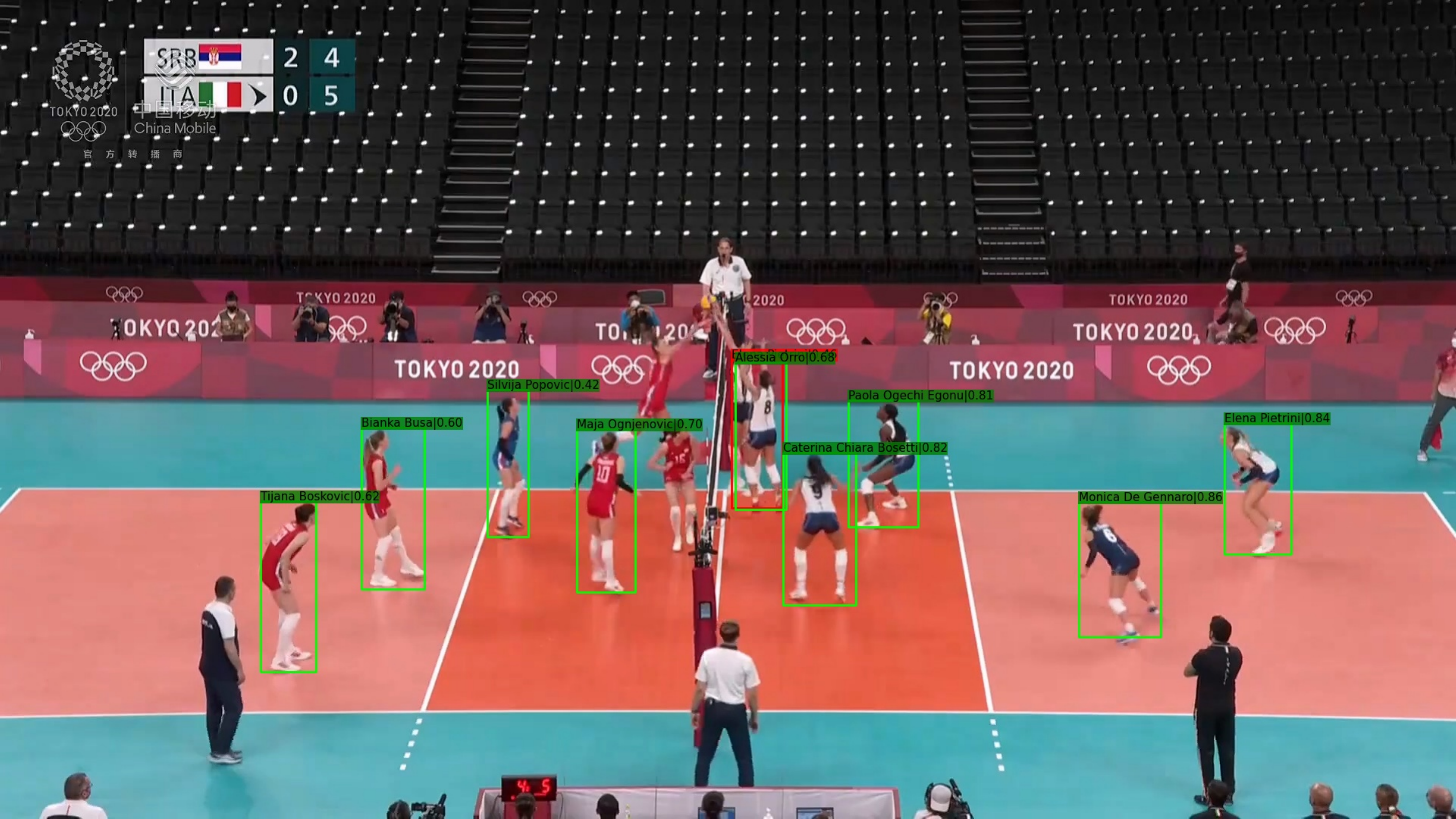}%
\label{fig_volleyball_4}}\vspace{-0.5em}
\caption{\small Visualization of experimental results on Volleyball-IOD benchmark: (a) Detection results with GFL training on all data; (b) Detection results with GFL by directly fine-tuning  (Catastrophic Forgetting); (c) Incremental results of ERD; (d) Incremental results of $R^2D$.}\vspace{-12pt}
\label{fig_volleyball}
\end{figure*}

Figure \ref{fig_volleyball} showcases the visualization results on Volleyball-IOD.
In Figures~\ref{fig_volleyball_2} and~\ref{fig_volleyball_4}, directly fine-tuning fails to detect players from previously learned tasks and only detects players from the current task.
Conversely, $R^2D$ detects players of the most recent task and recalls the knowledge about players in previous tasks. 
In Figure~\ref{fig_volleyball_3}, ERD presents a less satisfactory performance due to more detection errors in the objects.
Remarkably, our method performs closer to the Upper Bound method of Figure~\ref{fig_volleyball_1}. 
Nonetheless, all methods demonstrate limited performance near the center line, owing to the unavailability of features in the detection process caused by severe occlusion.
Synthesizing these observations, we conclude that $R^2D$ has superior robustness across various sports scenarios.


\subsection{Ablation Study}
We perform ablation studies to prove the effectiveness of each component. 
In Table~\ref{table_ablation}, we report results under three-step and five-step settings. 
In this table, \textbf{All} denotes the entire regions are distilled in IOD. 
\textbf{Cand} denotes only rough candidate regions are distilled in IOD.
\textbf{Refine} denotes only refined valuable regions are distilled in IOD.
Besides, we employ different distillation loss on classification and localization branches to scrutinize the effect of our method. 
\textbf{DCD} denotes the loss formulated in Equation \ref{equation_cls_loss_high} that decouples the maximum and non-maximum classification responses.
\textbf{L1} denotes L1 loss in Equation \ref{equation_cls_loss_low}.
\textbf{LD} denotes LD loss with the normal temperature $\mathcal{T}_{1}$ following Equation \ref{equation_reg_loss_high}.
\textbf{LDLT} denotes LD loss with the lower temperature $\mathcal{T}_{2}$ following Equation \ref{equation_reg_loss_low}.
``+" sign in Table~\ref{table_ablation} denotes the combination of distillation methods.
For example, ``DCD+L1" denote that \textbf{DCD} loss is adopted for distilling high-value regions, and \textbf{L1} loss is adopted for distilling low-value regions.

\begin{table*}[t]
\centering
\caption{Incremental results ($AP$,\%) of ablation study on NBA-IOD under Five and Three Steps settings.}
\label{table_ablation}
\begin{tabular}{@{}cll|dg|dgef}
\toprule
\multicolumn{3}{c|}{Methods}                                                                   & \multicolumn{2}{c|}{Three-step}         & \multicolumn{4}{c}{Five-step}                                \\ 
\multicolumn{1}{c}{abbr}     & \multicolumn{1}{c}{classification}       & \multicolumn{1}{c|}{regression}& Step 2         & Step 3      & Step 2        & Step 3        & Step 4        & Step 5        \\ \midrule
\multicolumn{1}{c}{A1}       & \multicolumn{1}{l}{{\bf All}: DCD}       & {\bf All}: LD                  & 51.3           & 47.5         & 39.6          & 57.4          & 35.7          & 38.3          \\
\multicolumn{1}{c}{A2}       & \multicolumn{1}{l}{{\bf All}: L1}        & {\bf All}: LD                  & 57.4           & 51.3         & 63.6          & 66.0          & 42.8          & 45.4          \\
\multicolumn{1}{c}{A3}       & \multicolumn{1}{l}{{\bf Cand}: DCD}      & {\bf Cand}: LD                 & 73.2           & 63.7         & 68.6          & 69.5          & 56.8          & 58.5          \\
\multicolumn{1}{c}{A4}       & \multicolumn{1}{l}{{\bf Cand}: L1}       & {\bf Cand}: LD                 & 74.6           & 67.8         & 73.5          & 71.6          & 65.5          & 63.9          \\
\multicolumn{1}{c}{A5}       & \multicolumn{1}{l}{{\bf Refine}: DCD+L1} & {\bf Cand}: LD                 & \textbf{76.7}  & 70.4         & \textbf{75.3} & 73.2          & 67.7          & 65.5          \\
\multicolumn{1}{c}{A6}       & \multicolumn{1}{l}{{\bf Cand}: L1}       & {\bf Refine}: LD+LDLT          & 74.4           & 67.2         & 73.3          & 71.5          & 65.0          & 64.0          \\
\multicolumn{1}{c}{A7($R^2D$)} & \multicolumn{1}{l}{{\bf Refine}: DCD+L1} & {\bf Refine}: LD+LDLT          & 76.4           & \textbf{70.9}& 75.1          & \textbf{73.4} & \textbf{68.3} & \textbf{66.2} \\ \bottomrule
\end{tabular}
\end{table*}

{\bf {Rough Candidate Regions.}} 
In Table~\ref{table_ablation}, we conduct experiments A1, A2, A3, and A4 to analyze the effectiveness of candidate regions. 
Results show that A3 and A4, which involve incremental distillation on the candidate regions, perform significantly better than A1 and A2 in both three-step and five-step settings.
Moreover, by comparing A2 to A1 (or A4 to A3), we observe a clear boost when switching from DCD loss to L1 loss, and either A3 or A4 performs better than A2 and A1 on all incremental steps.
These findings demonstrate that incremental distillation on the candidate regions is more effective than on all regions and is robust to different classification distillation methods.
The rough candidate regions filter out ``bad" nodes with low confidence and redundant noise, allowing the distillation method to focus more on knowledge transfer during the incremental process.

{\bf {Refined Classification Distillation.}}
Table~\ref{table_ablation} presents experimental groups A1, A2, A3, A4, and A5, which were designed to investigate the effectiveness of refined classification distillation. 
We observe that employing the L1 loss results in better performance than the DCD loss, irrespective of whether the distillation region is introduced (A2 and A4).
Although L1 performs better than DCD, we believe it is prone to be less efficient than DCD in transferring foreground rich knowledge, as it treats different elements in the classification response indiscriminately, which may result in the learning of target categories being submerged in other non-target categories’ learning.
Accordingly, we propose a combined approach utilizing both DCD loss and L1 loss.
Specifically, we apply DCD loss to distill regions rich in foreground class knowledge and L1 loss to distill regions predominantly associated with the background.
As high-value regions tend to represent foreground classes and low-value regions typically correspond to background classes, we empirically employ DCD loss for high-value regions and L1 loss for low-value regions.
In group A5, the combination of DCD and L1 outperforms the use of either loss function alone in all incremental steps, corroborating our previous analysis.

{\bf {Refined  Distillation in Classification and Regression Heads.}} 
Table~\ref{table_ablation} presents experimental groups A4, A5, A6, and A7, which were designed to evaluate the effectiveness of applying refined distillation to both classification and regression heads.
Compared to A4, A6 shows a slight improvement in the fifth step of the five-step setting, but underperforms in other incremental steps and the three-step setting. This suggests that refining distillation on the regression head alone does not consistently improve performance.
In contrast, when comparing A5 and A7, although A7 marginally trails A5 in the second step of both five-step and three-step settings, it outperforms A5 in the remaining incremental steps and overall. 
These findings indicate that the best performance gains are achieved by applying refined distillation simultaneously to both classification and regression heads, as opposed to using it in a single branch.
The slightly lower performance of A7 compared to A5 in the second step of both settings can be attributed to the properties of LDLT.
LDLT employs a lower distillation temperature, focusing more on the knowledge in larger elements and less on that in smaller elements across the regression response.
During the early steps, the low-value regions of the regression head have a relatively lower noise presentation. 
The performance gains provided by LDLT become increasingly evident with additional incremental steps, as knowledge in smaller elements is gradually forgotten. As shown in Table~\ref{table_ablation}, our proposed method improves upon A5 by 0.2\%, 0.6\%, and 0.7\% in step 3, 4, and 5 of the five-step setting, respectively. 
This progressive increase in accuracy gain supports our hypothesis.
Overall, these results demonstrate the superiority of our proposed refined response distillation method.

\subsection{Discussion}
In this section, we will discuss the hyperparameters $\theta$ and $\mathcal{T}_2$, and provide a further perspective on refined regions and superior performance.
\begin{table}[h]
\centering
\caption{Incremental results($AP$, \%) under different $\theta$.}
\label{table_abla_thela}
\begin{tabular}{@{}l|dg|dgef}
\toprule
\multirow{2}{*}{$\theta$} & \multicolumn{2}{c|}{Three-step}  & \multicolumn{4}{c}{Five-step}                           \\ 
                   & Step 2         & Step 3         & Step 2        & Step 3        & Step 4        & Step 5          \\ \midrule
0                  & 62.7           & 55.3           & 66.8          & 67.6          & 44.3          & 50.1            \\
0.05               & \textbf{76.4}  & \textbf{70.9}  & \textbf{75.1} & \textbf{73.4} & \textbf{68.3} & 66.2            \\
0.1                & 76.1           & 70.7           & 74.3          & 72.6          & 67.0          & 65.7            \\
0.2                & 74.5           & 69.0           & 74.1          & 73.1          & 68.0          & \textbf{66.3}   \\
0.3                & 74.3           & 67.9           & 73.1          & 72.7          & 67.0          & 65.9            \\
0.4                & 73.9           & 67.3           & 73.3          & 72.7          & 66.6          & 65.6            \\ \bottomrule
\end{tabular}
\end{table}

{\bf{Threshold $\theta$ for Rough Candidate Regions.}} 
To assess the effectiveness of the hyper-parameter $\theta$ as a threshold for selecting rough candidate regions in Equation \ref{equation_cand_region}, we conduct experiments with $\theta$ ranging from 0 to 0.4. 
The results are presented in Table~\ref{table_abla_thela}.
Notably, when $\theta=0$, all regions are selected as the candidate regions.
Table~\ref{table_abla_thela} shows that $\theta=0.05$ yields better results than $\theta$ values of 0, 0.1, 0.3, and 0.4 under both settings. 
Furthermore, $\theta =0.05$ performs slightly worse than $\theta =0.2$ in the fifth step of the five-step setting, while it performed better in other steps and in the three-step setting.
In conclusion, performance declines when $\theta$ is either too large or too small, and $\theta=0.05$ proves to be a robust threshold for the candidate region and outperforms other tested $\theta$ values.

\begin{table}[h]
\centering
\caption{Incremental results($AP$, \%) under different $\mathcal{T}_2$.}
\label{table_abla_t2}
\resizebox{0.48\textwidth}{!}{%
\begin{tabular}{@{}c|dg|dgef|d}
\toprule
\multirow{2}{*}{$\mathcal{T}_2$} & \multicolumn{2}{c|}{Three-step} & \multicolumn{4}{c|}{Five-step}                               & \multicolumn{1}{c@{}}{Home-and-away}         \\ 
                                 & Step 2         & Step 3           & Step 2        & Step 3        & Step 4        & Step 5        & Step 2                                 \\ \midrule
1                                & 76.4           & 70.4             & 75.3          & 73.8          & 67.9          & 65.6          & \textbf{72.4}                          \\
3                                & \textbf{76.6}  & 70.6             & 74.9          & 73.1          & 66.8          & 65.1          & 71.8                                   \\
5                                & 76.4           & \textbf{70.9}    & 75.1          & 73.4          & \textbf{68.3} & \textbf{66.2} & 72.3                                   \\
7                                & 76.5           & 70.5             & \textbf{75.4} & 73.4          & 67.9          & \textbf{66.2} & 71.9                                   \\
9                                & \textbf{76.6}  & 70.7             & 75.3          & \textbf{74.0} & \textbf{68.3} & 65.7          & 71.2                                   \\ \bottomrule
\end{tabular}
}\end{table}

\begin{figure*}[thb]
\centering
\subfloat[\small Classification Quality]{\includegraphics[width=0.24\textwidth]{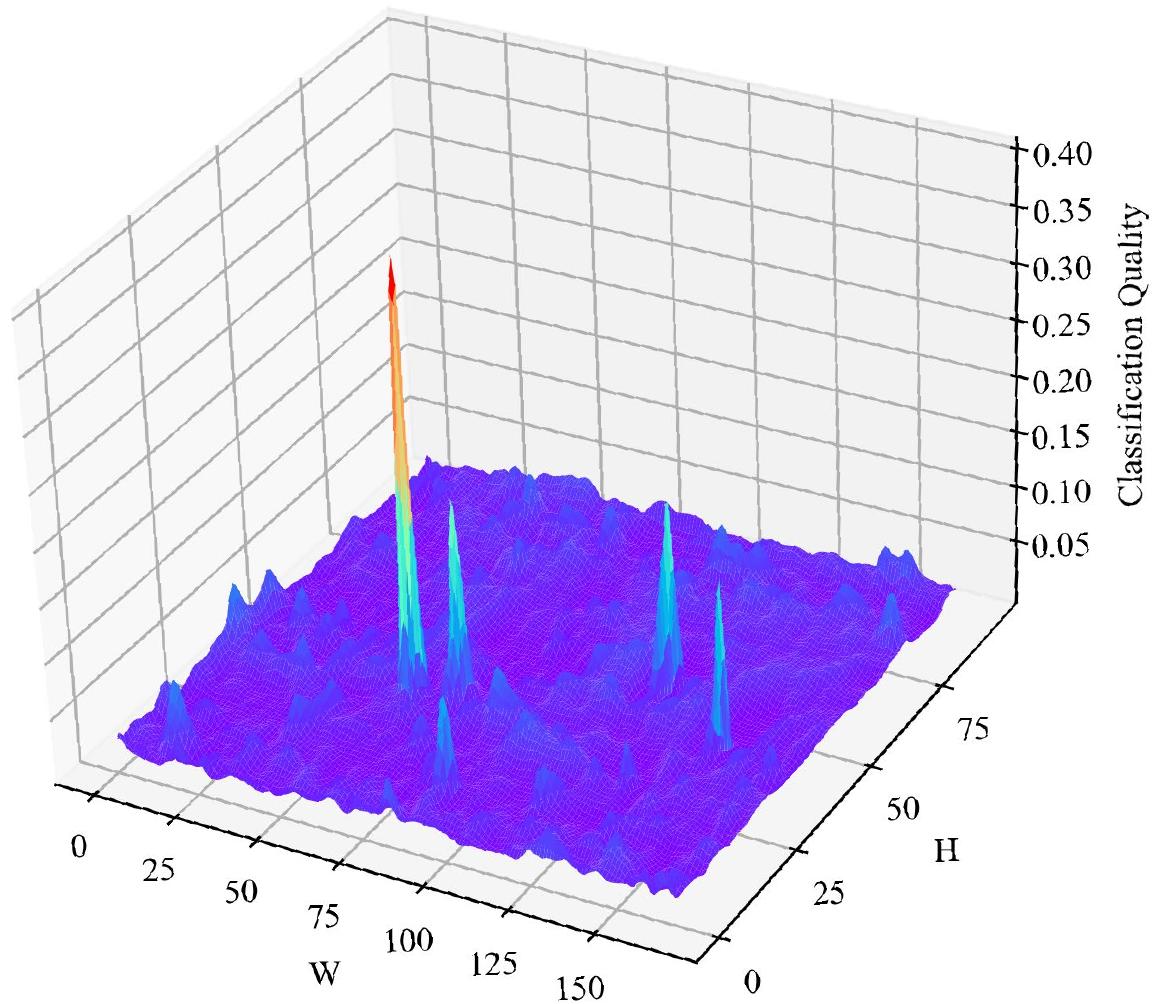}%
\label{fig_region_cls_all}}
\hfil
\subfloat[\small Candidate Regions]{\includegraphics[width=0.24\textwidth]{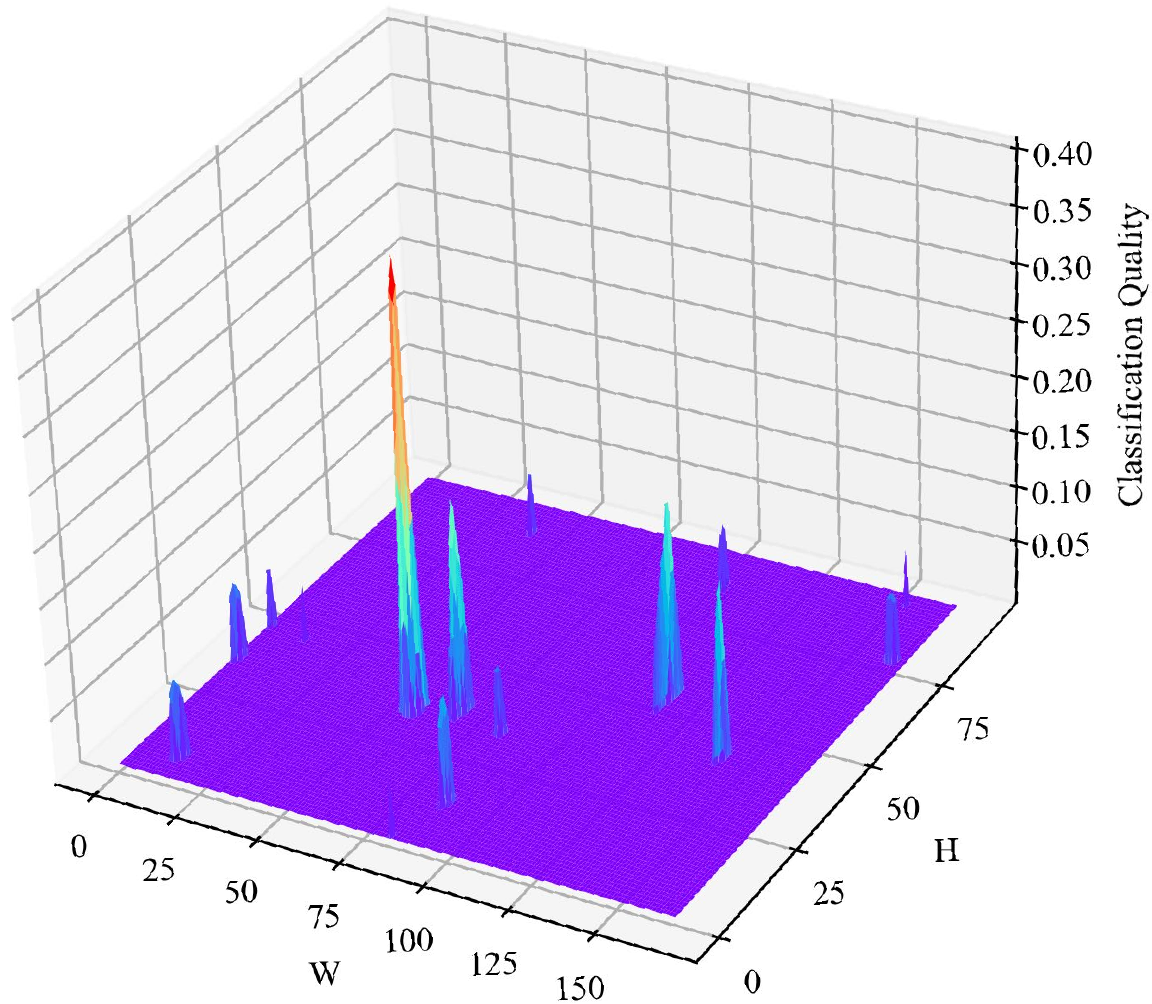}%
\label{fig_region_cls_cand}}
\hfil
\subfloat[\small High-Value Regions]{\includegraphics[width=0.24\textwidth]{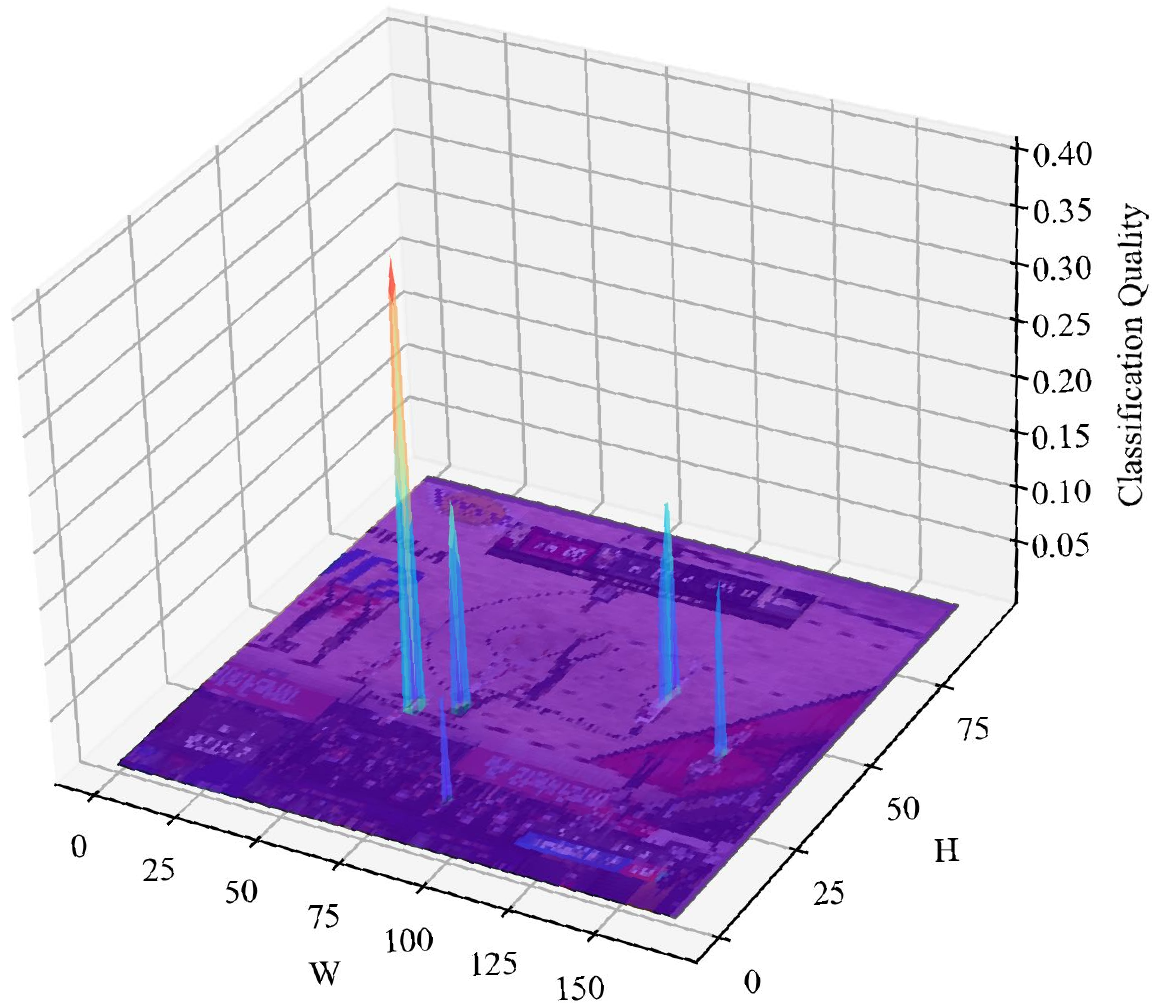}%
\label{fig_region_cls_high}}
\hfil
\subfloat[\small Low-Value Regions]{\includegraphics[width=0.24\textwidth]{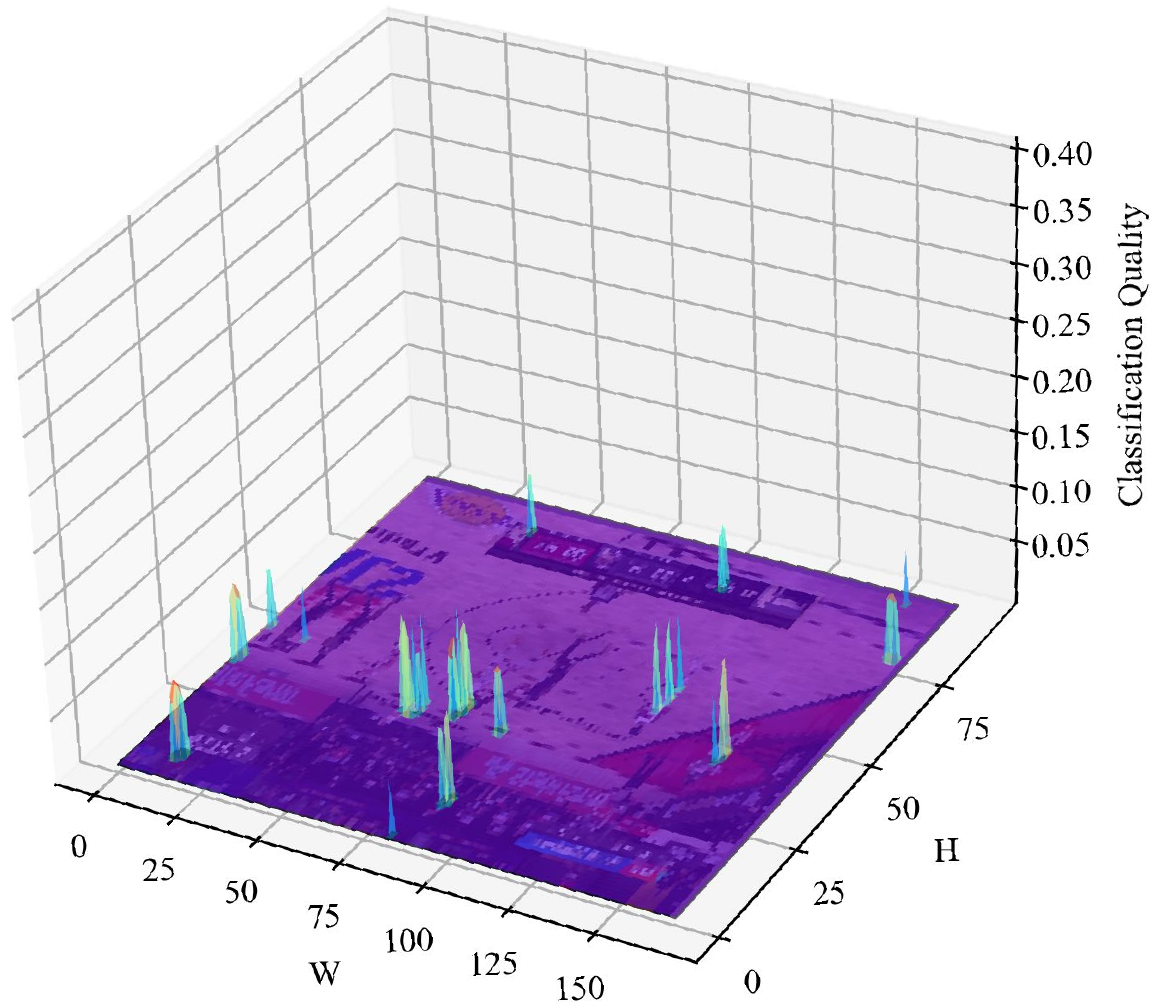}%
\label{fig_region_cls_low}}\\\vspace{-1em}
\subfloat[\small Localization Quality]{\includegraphics[width=0.24\textwidth]{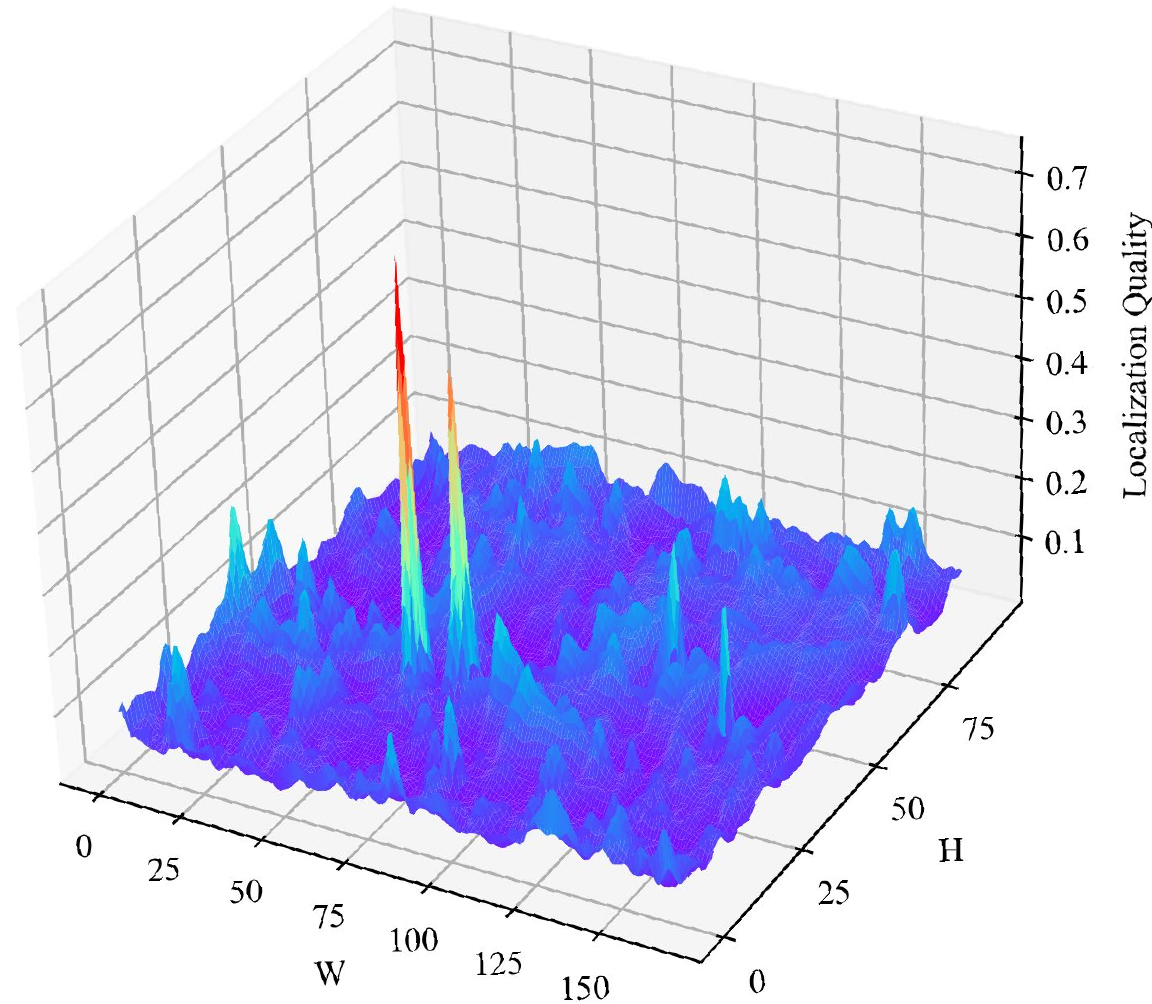}%
\label{fig_region_reg_all}}
\hfil
\subfloat[\small Candidate Regions]{\includegraphics[width=0.24\textwidth]{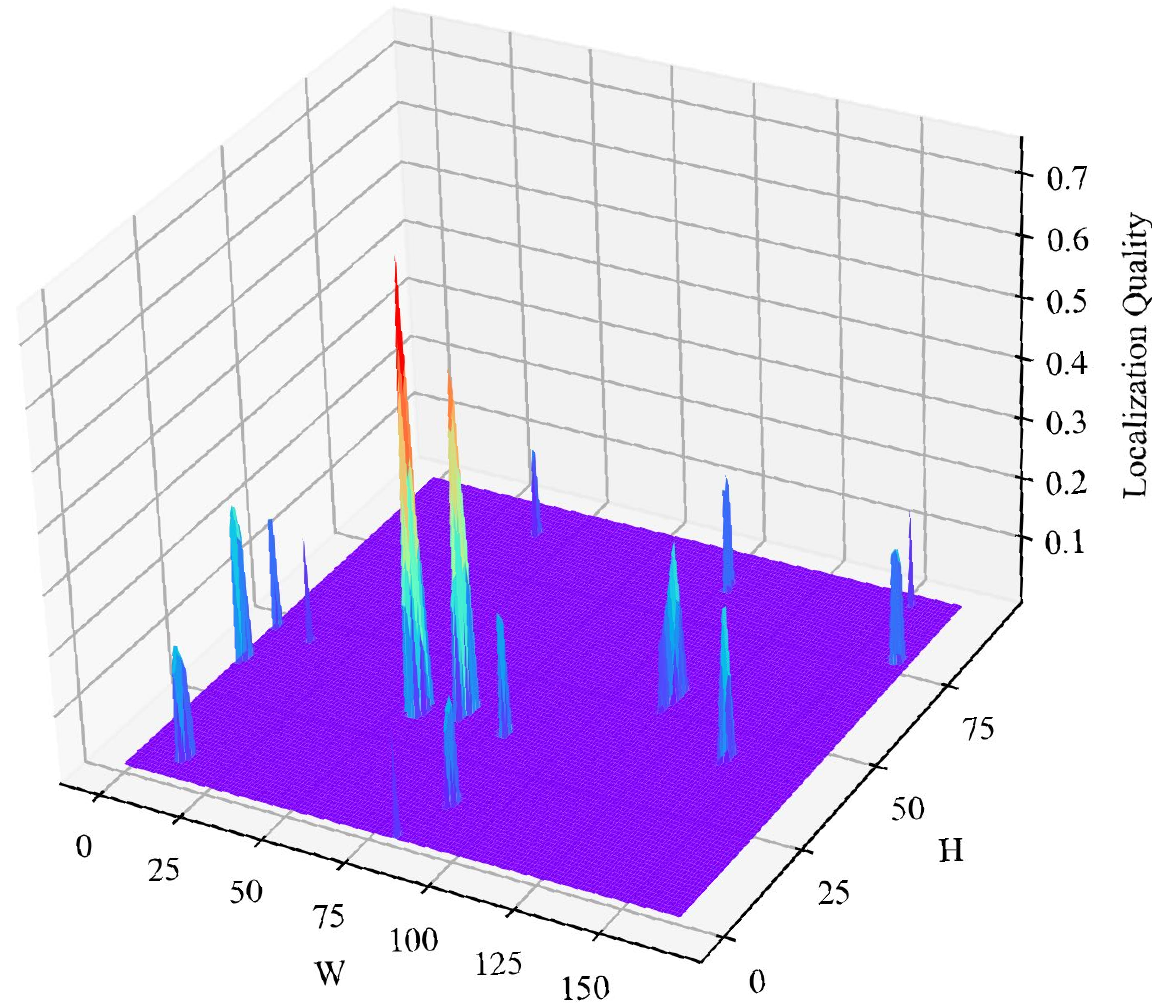}%
\label{fig_region_reg_cand}}
\hfil
\subfloat[\small High-Value Regions]{\includegraphics[width=0.24\textwidth]{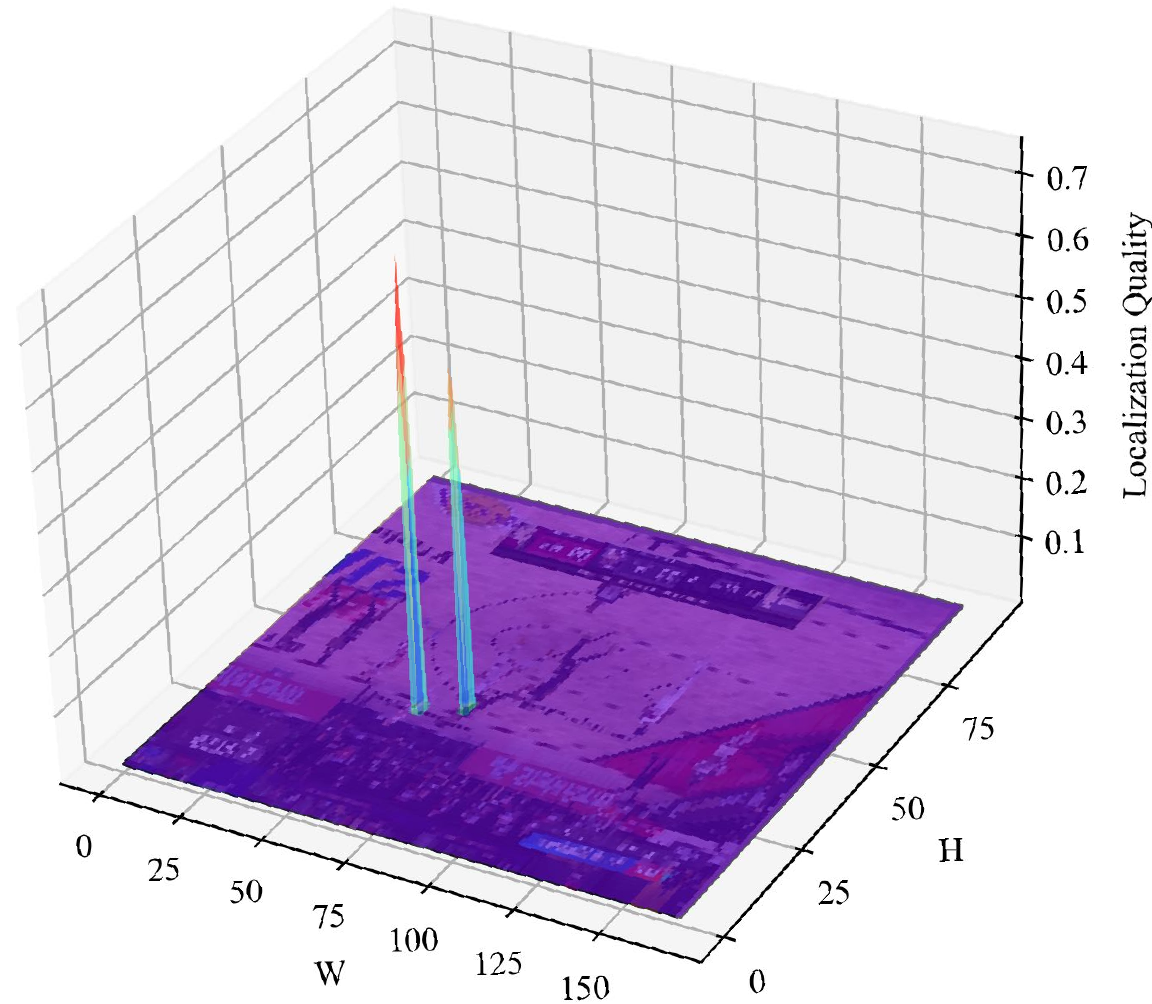}%
\label{fig_region_reg_high}}
\hfil
\subfloat[\small Low-Value Regions]{\includegraphics[width=0.24\textwidth]{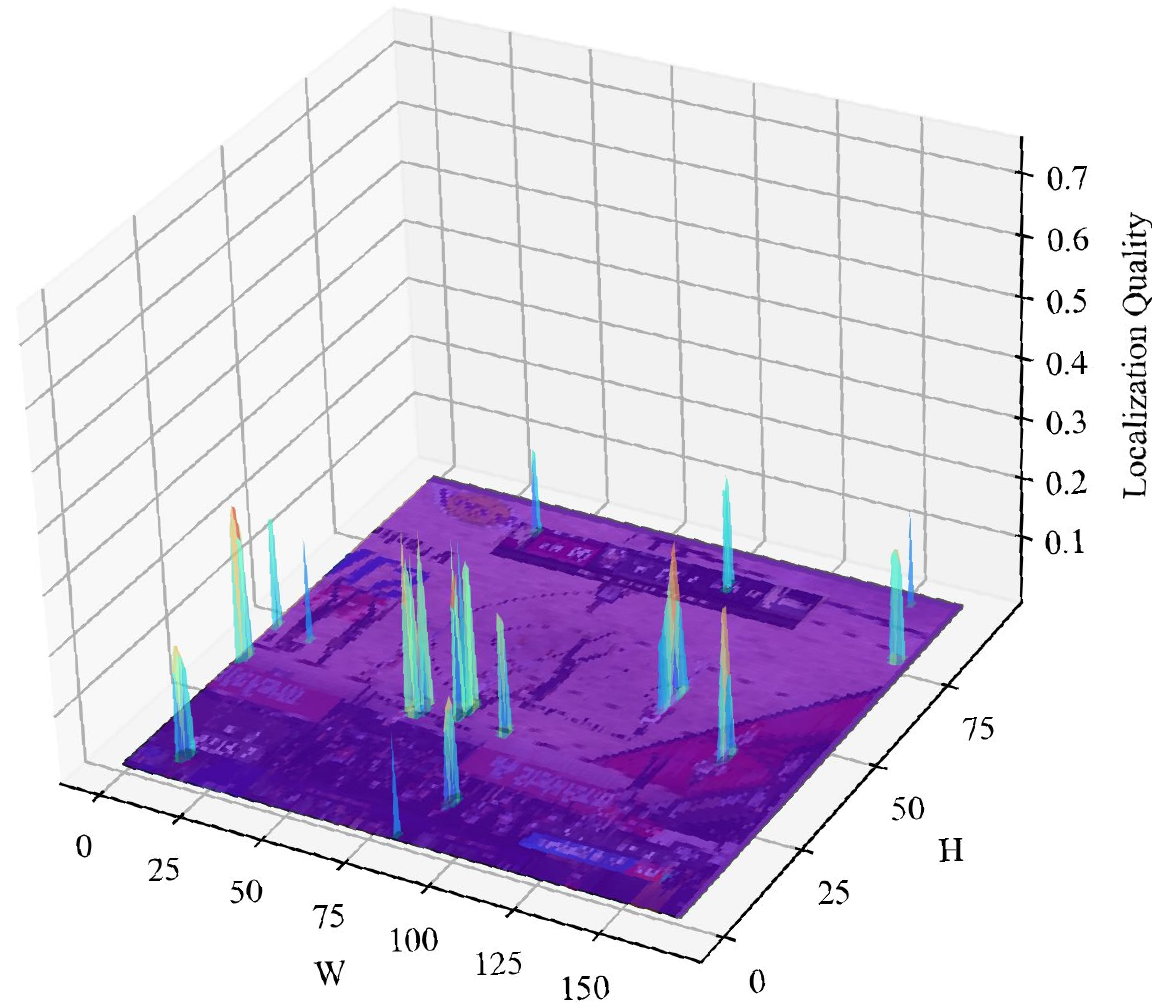}%
\label{fig_region_reg_low}}\vspace{-0.5em}
\caption{\small Visualization of refined distillation regions in classification and regression responses (P3 level of FPN). (a), (b), (c), and (d) denotes classification response quality for the all regions, candidate regions, high-value regions, and low-value regions, respectively. (e), (f), (g), and (h) denotes regression response quality in the corresponding regions. }\vspace{-12pt}
\label{fig_region}
\end{figure*}

{\bf{Temperature $\mathcal{T}_{2}$ for Localization Distillation.}} 
We perform five experiments to investigate the robustness of the hyper-parameter $\mathcal{T}_{2}$, which determines the distillation temperature for the low-value region in the regression head. 
Our results are summarized in Table~\ref{table_abla_t2}. 
In the three-step setting, $\mathcal{T}_{2}=5$ achieves the best result of 70.9\% in step 3. 
Although the accuracies of $\mathcal{T}_{2}=5$ and $\mathcal{T}_{2}=1$ in step 2 are the lowest at 76.4\%, the performance is only marginally impacted by different $\mathcal{T}_{2}$ values.
In the five-step setting, $\mathcal{T}_{2}=5$ attains the best results in steps 4 and 5, with slightly lower performance in steps 2 and 3. 
In the home-and-away setting, $\mathcal{T}_{2}=5$ achieves an accuracy of 72.3\%, which is just below the highest value of 72.4\% obtained with $\mathcal{T}_{2}=9$. Through comprehensive comparisons across various incremental scenarios, we observe that $\mathcal{T}_{2}$ exhibits different sensitivities to different tasks. 
This is because changes in task types introduce distinct features, leading to variations in performance.
Although $\mathcal{T}_{2}=5$ does not achieve the highest accuracy in every step across all scenarios, it generally exhibits balanced performance.
Therefore, we consider $\mathcal{T}_{2}=5$ a robust choice for diverse scenarios in this paper.

{\bf{Visualization of Refined Distillation regions.}} 
We present visualizations of the refined distillation regions in the P3 level (the first layer of the FPN) of the classification and regression head in Figure \ref{fig_region}.
Figures \ref{fig_region_cls_all} and \ref{fig_region_reg_all} show the raw distillation quality of responses, while Figures \ref{fig_region_cls_cand} and \ref{fig_region_reg_cand} filter out low-quality response nodes, forming a more refined candidate regions.
Compared to Figure \ref{fig_region_cls_cand}, Figures \ref{fig_region_cls_high} and \ref{fig_region_cls_low} have more precise classification distillation regions, while Figures \ref{fig_region_reg_high} and \ref{fig_region_reg_low} have more precise localization distillation regions than Figure \ref{fig_region_reg_cand}.
This indicates that the proposed method can maintain the classification and localization knowledge from previous tasks in a refined way.
Besides, comparing Figures \ref{fig_region_cls_high} and \ref{fig_region_cls_low}, we observe that the high-value regions in the classification branch are more likely to present the foreground classes, while the low-value regions are more likely to present the background. By comparing Figures \ref{fig_region_reg_high} and \ref{fig_region_reg_low}, we observe that the high-value regions in the regression branch have a higher certainty of existing players, while the low-value regions contain more noise.
Notably, the proposed method achieves a refined incremental detection method through the separation of high-value and low-value distillation regions. These results support our motivation for the proposed $R^2D$.

{\bf{Further Perspective on Superior Performance.}} In Figures \ref{fig_three}, \ref{fig_five}, \ref{fig_home_and_away}, \ref{fig_two_ranks}, and \ref{fig_volleyball}, we observe that existing incremental detection schemes maintain some learned knowledge about the players, however, they retain this knowledge while further ensuring the accuracy of the knowledge is difficult. In sports broadcast scenarios, the primary challenge stems from the pronounced feature homogeneity resulting from identical team uniforms, this challenge further introduces the IOD tasks into trouble. In this paper, the proposed $R^2D$ method provides insights for maintaining learned knowledge about the players while further ensuring accuracy.

\section{ Conclusions}
In this paper, we propose the $R^2D$ method to effectively mitigate catastrophic forgetting in the context of incremental player detection tasks. 
Subsequently, we introduce the NBA-IOD and Volleyball-IOD datasets as benchmarks for these tasks. 
Then, we conduct a systematic study of class incremental detection for players, utilizing these benchmarks.
Our comprehensive experiments reveal that $R^2D$ successfully retains learned knowledge about the players while concurrently preserving the accuracy of this knowledge.
In the future, we will study the refined incremental distillation on more excellent detectors (e.g., Deformable DETR~\cite{refx43}) to improve the knowledge transfer capabilities for players' IOD.
Extending our methodology to other sports activities for analyzing players' characteristics is also an important research direction.

\end{document}